%% file: main.tex
\documentclass{defaltmlr}
\input{math_commands.tex}

\usepackage{url}
\usepackage[table]{xcolor}
\usepackage[dvipsnames]{xcolor}
\usepackage{graphicx}
\usepackage{float}
\usepackage{geometry}
\usepackage{booktabs}
\usepackage[T1]{fontenc}
\usepackage{fontawesome}
\usepackage{fancyhdr}
\usepackage{xspace}
\usepackage{enumitem}
\usepackage{etoc}
\usepackage{titletoc}
\usepackage{tcolorbox}
\usepackage{listings}

\usepackage{algorithm}
\usepackage{algpseudocode}

\usepackage{color}
\usepackage[dvipsnames]{xcolor}
\definecolor{JalapenoRed}{RGB}{183,21,64}
\definecolor{Belize}{RGB}{41,128,185}
\definecolor{Amour}{RGB}{238,82,83}
\usepackage{colortbl}

\usepackage{hyperref}

\usepackage[most]{tcolorbox}
\usepackage{enumitem}

\tcbset{
  promptbox/.style={
    title={#1},
    width=\columnwidth,
    breakable,
    enhanced,
    colback=gray!3,
    colframe=black!30,
    coltitle=black,
    fonttitle=\bfseries\footnotesize,
    boxrule=0.8pt,
    arc=1mm,
    boxsep=2pt,
    left=4pt,
    right=4pt,
    top=4pt,
    bottom=4pt,
    before skip=4pt,
    after skip=6pt
  }
}

\newcolumntype{L}[1]{>{\raggedright\arraybackslash}p{#1}}

\usepackage{titlesec}
\usepackage{titletoc}
\usepackage[most]{tcolorbox}
\usepackage{enumitem}
   
\usepackage{tocloft}

\usepackage[table]{xcolor}

\definecolor{GroundBG}{HTML}{6e8eb7}
\definecolor{ProfileBG}{HTML}{d79a77}
\definecolor{InteractBG}{HTML}{d77434}
\definecolor{OursBG}{HTML}{3e9b4e}

\usepackage[T1]{fontenc}
\usepackage[utf8]{inputenc}
\usepackage{microtype}
\usepackage{inconsolata}
\usepackage{graphicx}

\usepackage{tabularx}

\usepackage[misc]{ifsym}
\usepackage{wrapfig}

\hypersetup{linkcolor=cyan}
\usepackage{mathtools}
\usepackage{ulem}
\usepackage{pifont}
\usepackage{bbding}

\usepackage{tcolorbox}
\usepackage{makecell}

\makeatletter
\let\orig@fnsymbol\@fnsymbol
\def\@fnsymbol#1{\ifcase#1\or\relax\else\orig@fnsymbol{#1}\fi}
\makeatother

\usepackage{tcolorbox}
\usepackage{enumitem}
\input{appendix/domain_style}

\tcbset{promptstyle/.style={%
    colback=blue!5!white, colframe=blue!40!black,
    arc=2pt, boxrule=0.5pt,
    left=6pt, right=6pt, top=4pt, bottom=4pt, boxsep=2pt,
    enhanced jigsaw,
    before skip=5pt, after skip=6pt,
    fonttitle=\small\bfseries, fontupper=\small
}}
\newtcolorbox{promptbox}[1]{%
    promptstyle, breakable, title={#1}, title after break={#1 (continued)}}
\newtcblisting{prompttemplate}[1]{%
    promptstyle, title={#1}, title after break={#1 (continued)}, listing only,
    listing options={basicstyle=\footnotesize\rmfamily, columns=fullflexible,
        breaklines=true, breakatwhitespace=true, breakindent=0pt,
        keepspaces=false, showstringspaces=false, aboveskip=0pt, belowskip=0pt,
        escapeinside={(*@}{@*)}}
}

\newcommand{\greensquare}{
    \tikz[baseline=1.1ex]{
        \node[fill, green, opacity=0.1, rounded corners, minimum width=2.5ex, minimum height=2.2ex, anchor=center, shape=rectangle] at (1ex,2ex) {};
    }
}

\newcommand{\redsquare}{
    \tikz[baseline=1.1ex]{
        \node[fill, red, opacity=0.3, rounded corners, minimum width=3.ex, minimum height=2.2ex, anchor=center, shape=rectangle] at (1ex,2ex) {};
    }
}

\newcommand{\yellowsquare}{
    \tikz[baseline=1.1ex]{
        \node[fill, yellow, opacity=0.3, rounded corners, minimum width=3.ex, minimum height=2.2ex, anchor=center, shape=rectangle] at (1ex,2ex) {};
    }
}

\def\method{\texttt{Intent-OPSD}}
\def\benchmark{\texttt{Intent-Eval}}

\newcommand{\symbolimg}[2][0.3cm]{%
  \ensuremath{\vcenter{\hbox{\includegraphics[height=#1]{#2}}}}%
}

\newcommand{\llama}{\symbolimg[0.35cm]{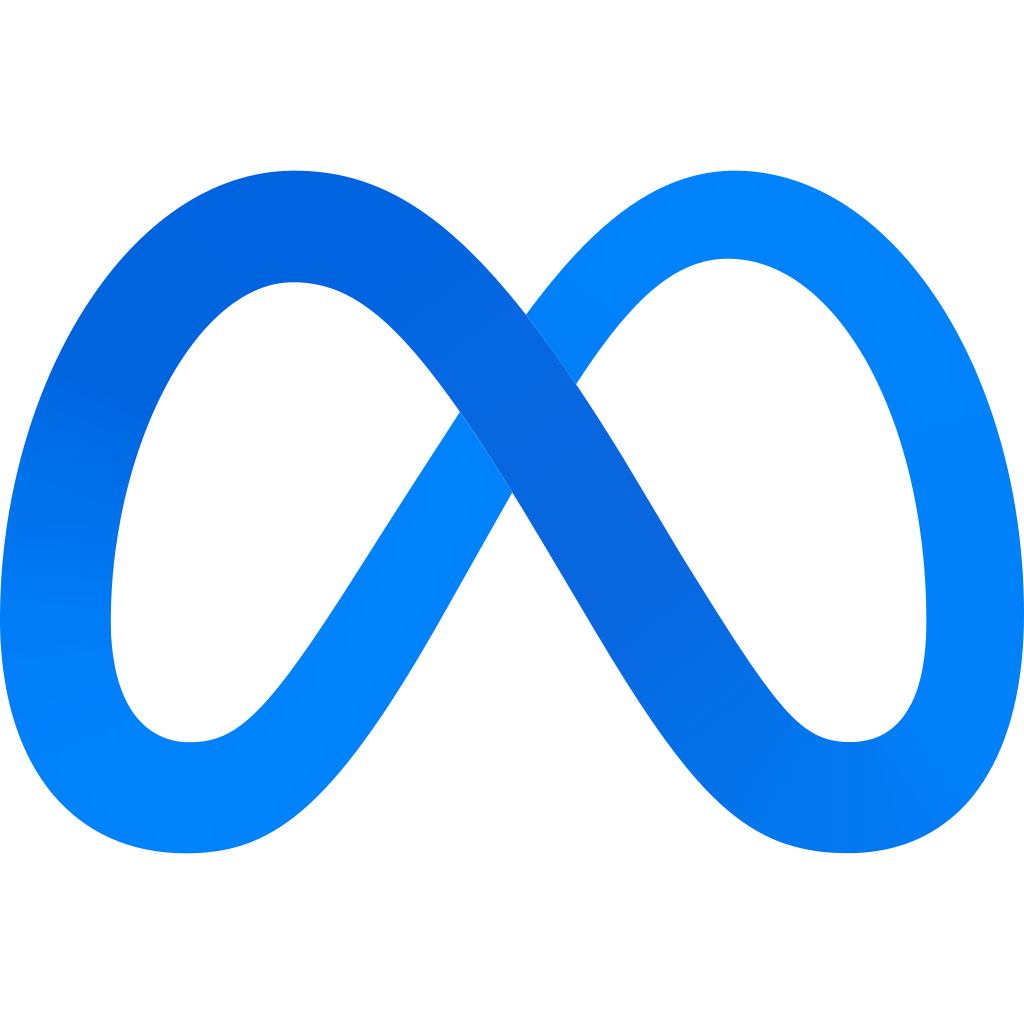}}
\newcommand{\qwen}{\symbolimg[0.35cm]{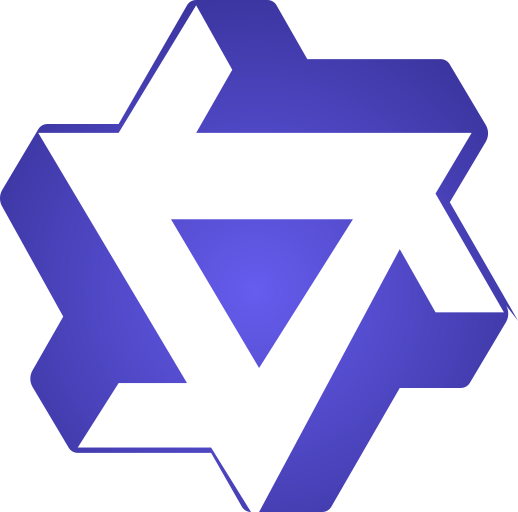}}
\newcommand{\gemini}{\symbolimg[0.35cm]{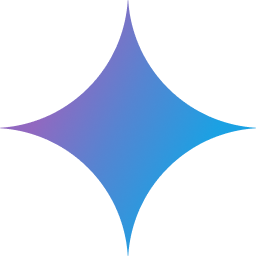}}
\newcommand{\claude}{\symbolimg[0.35cm]{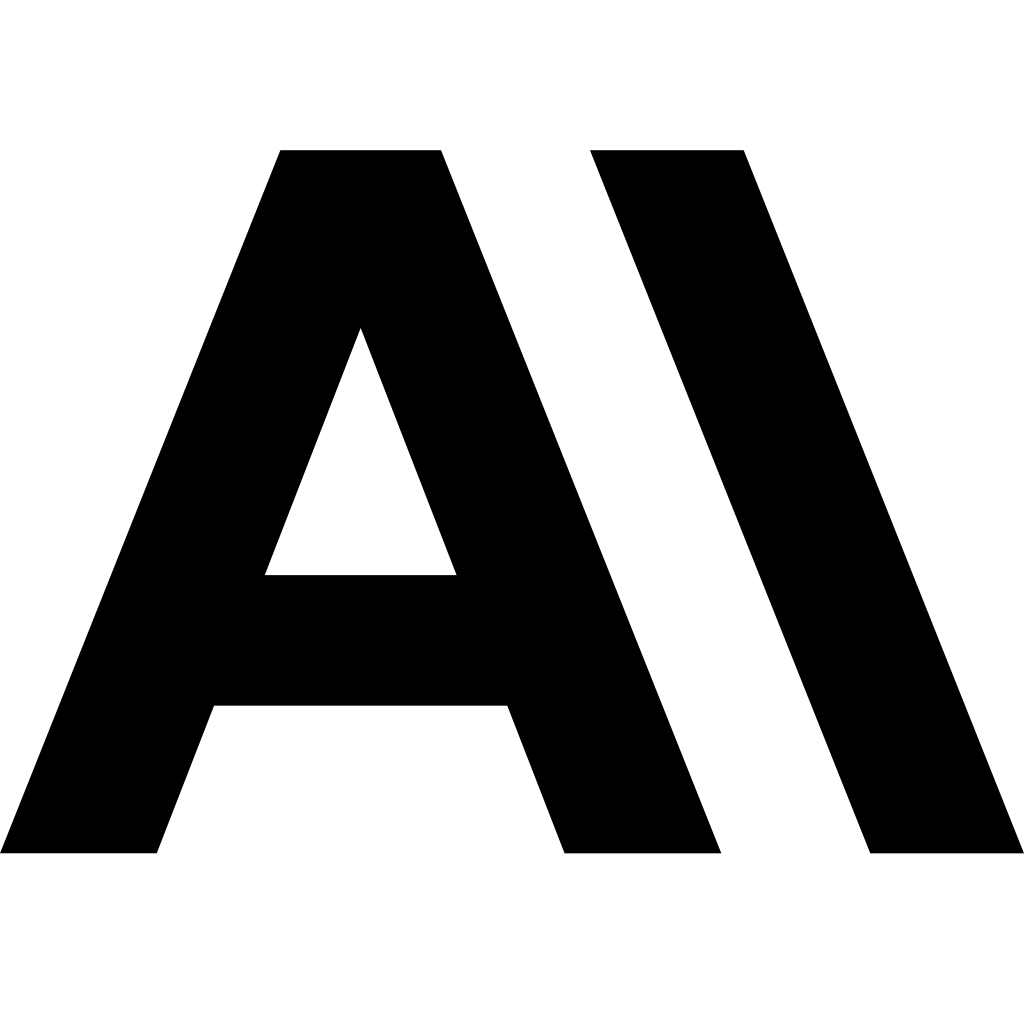}}
\newcommand{\openai}{\symbolimg[0.35cm]{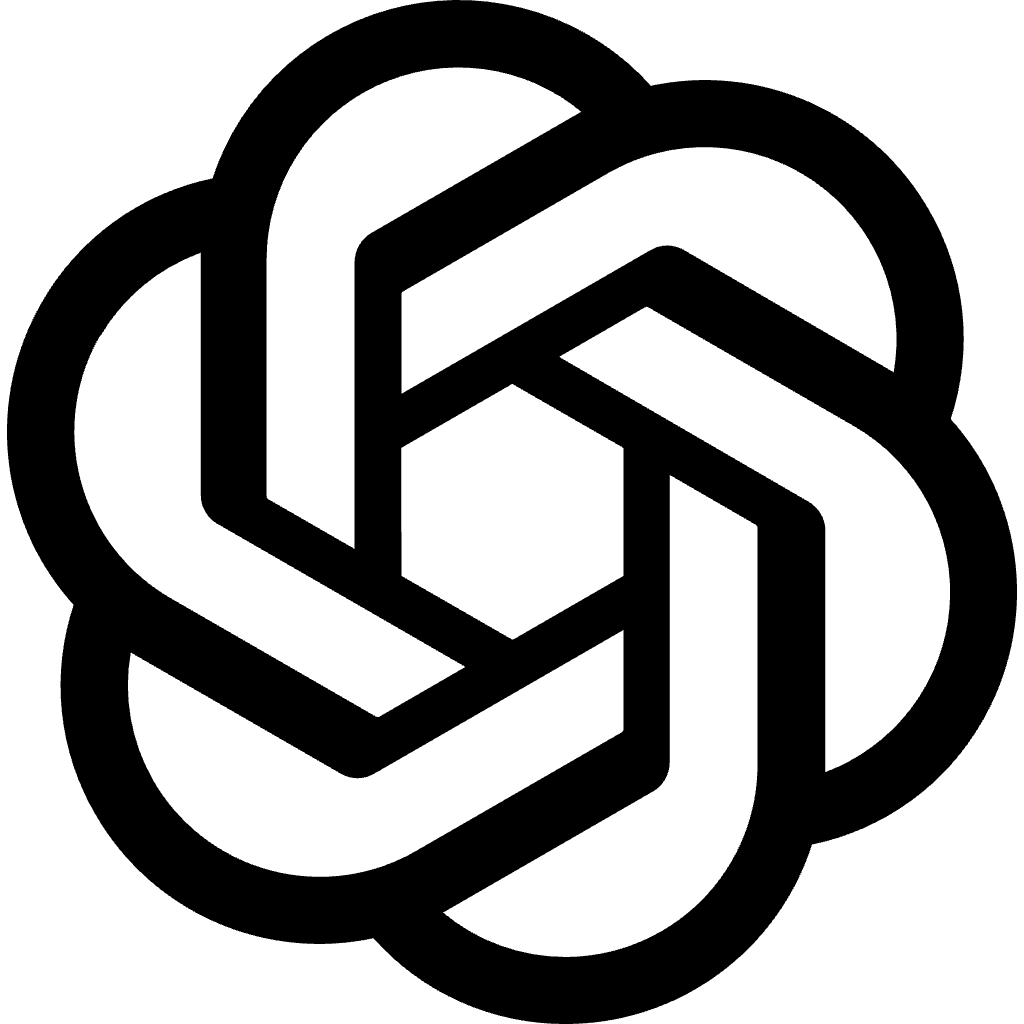}}
\newcommand{\deepseek}{\symbolimg[0.35cm]{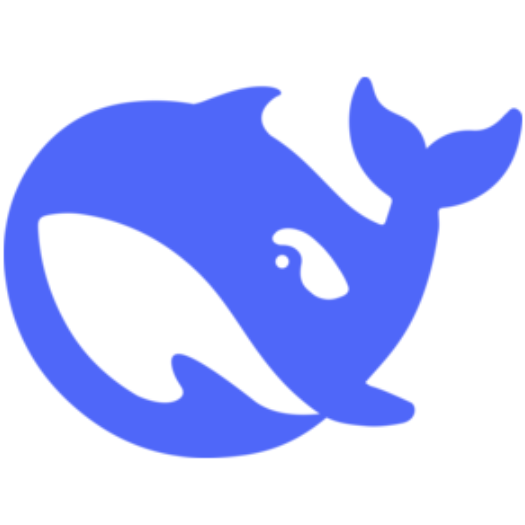}}

\newcommand{\papertitle}{You Changed Your Mind, The Model Didn’t: Demystifying Intent in Multi-Turn Dialogue}
\title{\papertitle}
\hypersetup{pdftitle={\papertitle}}

\author{
\parbox{\textwidth}{
Junle Chen\textsuperscript{1}, \hspace{0.5mm}
Wei Chen\textsuperscript{1,2$~^{\textrm{\Letter}}$}, \hspace{0.5mm}\\
\textbf{Zhengjun Huang}\textsuperscript{1}, \hspace{0.5mm}
\textbf{Zhoujin Tian}\textsuperscript{1}, \hspace{0.5mm}
\textbf{Yuxuan Liu}\textsuperscript{1}, \hspace{0.5mm}
\textbf{Kai Wang}\textsuperscript{2}, \hspace{0.5mm}
\textbf{Rui Chen}\textsuperscript{2}, \hspace{0.5mm}
\textbf{Xiaofang Zhou}\textsuperscript{1} \hspace{0.5mm}
}
}

\affiliation{\textsuperscript{1}HKUST, \hspace{0.5mm} \textsuperscript{2}Tencent Hy}

\abstract{
When a large language model handles a multi-turn task and a user proposes a change but ultimately rejects it, the model should continue as if nothing changed. We find a surprising failure: merely mentioning a rejected change can derail task execution, even when the user's final intent remains unchanged. To systematically study language model behavior under evolving user intent, we introduce \benchmark{}, a controlled benchmark spanning tool actions, code, databases, and mathematics. Across diverse tasks, models are vulnerable to both rejected proposals and superseded requirements, consistent with \textit{mentioned-as-in-effect} confusion: conversational content is treated as active requirements even after it has been rejected or replaced. Accuracy degradation can deepen or persist as interaction continues, highlighting the need to distinguish what has been mentioned from what remains in effect. Building on this insight, we propose \method{}, a decision-conditioned on-policy self-distillation framework with Teacher and Student initialized from the same model. The frozen Teacher provides active-intent supervision from the complete task matching the user's decision, training the Student on the full dialogue to follow active requirements reflecting user intent.
}

\correspondence{\textcolor{myblue}{jchenkg@connect.ust.hk}, \textcolor{myblue}{onedeanxxx@gmail.com}}
\repository{\url{https://github.com/junle-chen/intent}}
\website{\url{https://junle-chen.github.io/intent-site/}}
\date{September 24, 2026}

\titlespacing*{\section}{0pt}{1.8ex plus 0.4ex minus 0.2ex}{0.8ex}
\titlespacing*{\subsection}{0pt}{1.4ex plus 0.3ex minus 0.2ex}{0.6ex}
\titlespacing*{\paragraph}{0pt}{0.8ex plus 0.2ex}{0.6em}
\begin{document}

\maketitle

\makeatletter
\let\@fnsymbol\orig@fnsymbol
\makeatother

\input{content/1_introduction}
\input{content/3_benchmark}

\input{content/4_method}
\input{content/5_experiments}
\FloatBarrier
\input{content/2_relatedWork}

\input{content/6_conclusion}
\label{main_text_end}

\FloatBarrier

\definecolor{textgray}{HTML}{6E6E73}
\makeatletter
\newcommand\applefootnote[1]{%
  \begingroup
  \renewcommand\thefootnote{}%
  \renewcommand\@makefntext[1]{\noindent##1}%
  \footnote{#1}%
  \addtocounter{footnote}{-1}%
  \endgroup
}
\makeatother

\bibliography{ref}
\bibliographystyle{plainnat}
\clearpage
\appendix

\input{appendix/0_appendix_list}

\end{document}

%% file: math_commands.tex
\usepackage{amsmath,amsfonts,bm}

\def\eqref#1{equation~\ref{#1}}

\def\1{\bm{1}}

\DeclareMathAlphabet{\mathsfit}{\encodingdefault}{\sfdefault}{m}{sl}
\SetMathAlphabet{\mathsfit}{bold}{\encodingdefault}{\sfdefault}{bx}{n}



%% file: appendix/domain_style.tex
\definecolor{domainActions}{HTML}{7F5D1C}
\definecolor{domainCode}{HTML}{315E75}
\definecolor{domainDatabase}{HTML}{3A5F4A}
\definecolor{domainMath}{HTML}{794660}

\definecolor{intentBlue}{HTML}{2655A5}
\definecolor{intentOrange}{HTML}{F57419}
\definecolor{intentRed}{HTML}{E65143}
\definecolor{intentGreen}{HTML}{6B9E2F}

\DeclareRobustCommand{\intentblue}[1]{\textcolor{intentBlue}{#1}}
\DeclareRobustCommand{\intentorange}[1]{\textcolor{intentOrange}{#1}}
\DeclareRobustCommand{\intentred}[1]{\textcolor{intentRed}{#1}}
\DeclareRobustCommand{\intentgreen}[1]{\textcolor{intentGreen}{#1}}

\DeclareRobustCommand{\domainname}[1]{\textcolor{domain#1}{\textbf{#1}}}
\tikzset{domain icon/.style={x=1em,y=1em,baseline=-0.30em,
  line width=0.055em,line cap=round,line join=round}}

\DeclareRobustCommand{\domainlabel}[1]{%
  \mbox{\csname domainicon#1\endcsname\hspace{0.35em}\domainname{#1}}}

%% file: content/1_introduction.tex
\section{Introduction}
\label{sec:introduction}


Large language models complete tasks in multi-turn dialogue, where user intent unfolds gradually. Even with all information provided, spreading a task across turns reduces performance relative to a single message~\citep{laban2026lost}. Yet users do more than reveal information: they clarify requirements and float changes they later accept or reject. The context then holds rejected or superseded content that should not shape the answer. Recent benchmarks examine requirement changes~\citep{tack2026evolving,wu2026specpath}, but do not compare clarifications, accepted changes, and rejected changes on the same tasks. We compare all three and test whether their effects grow or persist in longer dialogues, asking whether models distinguish what was \textit{said} from what remains \textit{in effect}.

\begin{figure}[t]
\centering
\includegraphics[width=\linewidth]{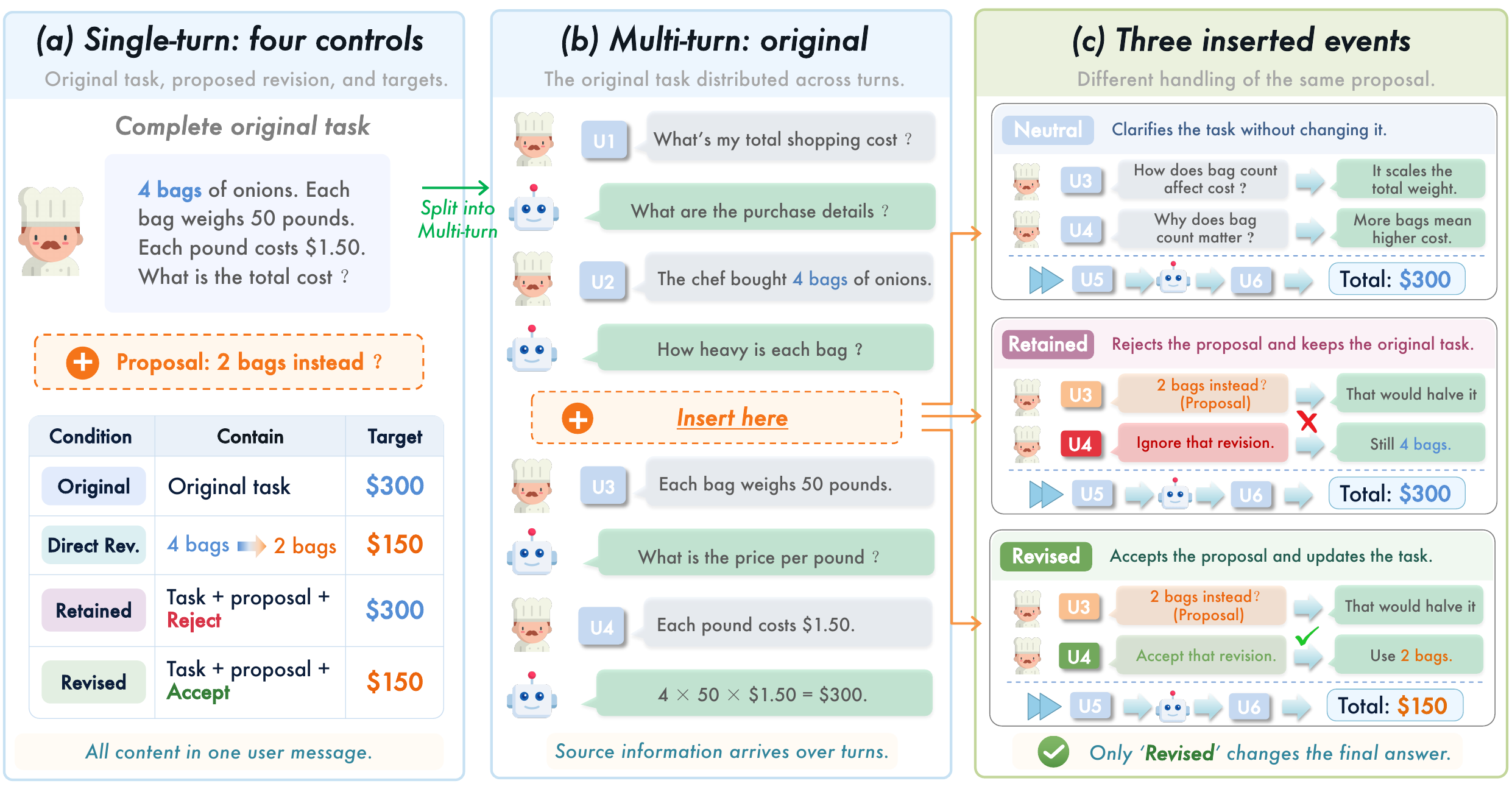}
\caption{Illustration of active intent.
\textbf{(a)}~Single-turn controls and targets.
\textbf{(b)}~The Original task distributed across turns; U3--U4 mark the insertion point for the other conditions.
\textbf{(c)}~Neutral clarifies; Retained rejects and Revised accepts the same proposal. Only Revised changes the target.}
\label{fig:construction}
\end{figure}

Figure~\ref{fig:construction} illustrates this distinction in a purchase calculation. Once the bag count ($4$ bags) is introduced, the dialogue splits into four branches: \textit{Original} continues without interruption, while \textit{Neutral} adds clarification without changing what is in effect. In both \textit{Retained} and \textit{Revised}, the user \textit{\textbf{mentions}} the same proposal ($2$ bags), rejecting it in \textit{Retained} and accepting it in \textit{Revised}. Only \textit{Revised} changes what remains \textit{\textbf{in effect}}, updating the answer from \$300 to \$150. Single-turn controls (Figure~\ref{fig:construction} (a)) state the same tasks in one message, separating multi-turn failures from task difficulty.

To systematically test whether models follow what remains in effect, we introduce \benchmark{}, a benchmark of 414 tasks across Actions, Code, Database, and Math ($3{,}312$ evaluation instances). Across eight models, we find three patterns of degradation:
\textit{\ding{182}~\textbf{Dispersed disclosure:}} splitting the task across turns lowers mean accuracy by $36.30$~pp from the single-turn control, even without interruptions (\textit{Original});
\textit{\ding{183}~\textbf{Conversational overhead:}} adding neutral clarification without changing the task (\textit{Neutral}) incurs an additional $4.43$~pp drop; and
\textit{\ding{184}~\textbf{Decision-induced confusion:}} a proposed change further degrades performance whether rejected or accepted, including in single-turn controls. In multi-turn dialogue, \textit{Retained} and \textit{Revised} average $8.06$ and $5.55$~pp below \textit{Original}, respectively.

We further analyze errors and performance in longer dialogues to understand these drops. Accuracy tends to fall as users ask for clarification over more turns or propose and reject more changes. Under \textit{Retained} and \textit{Revised}, new errors relative to \textit{Original} often contain rejected or superseded content, suggesting \textit{mentioned-as-in-effect} confusion: models treat what was said as still in effect.

To address this confusion, we use the model's stronger single-turn capability to help it follow what remains in effect in multi-turn dialogue. Notably, the original single-turn task shares active requirements with multi-turn \textit{Original}, \textit{Neutral}, and \textit{Retained}, while the revised single-turn task matches multi-turn \textit{Revised}. This correspondence motivates \method{}, which extends multi-turn self-distillation~\citep{chen2026found,zheng2026maigo} by matching the Teacher to the task that remains in effect after the user's decision. A frozen base model serves as the Teacher and receives the complete single-turn task matching the user's decision, while the Student, initialized from the same model, receives the full dialogue and is trained on-policy to match the Teacher. This focuses supervision on active requirements while the Student learns from the dialogue. At inference, the Student needs neither the Teacher nor the single-turn prompt.
Our main contributions are:

\begin{itemize}[leftmargin=*,topsep=0pt,partopsep=0pt,itemsep=2pt,parsep=0pt]
\item \textbf{\benchmark{}:} A four-domain benchmark with four controlled
multi-turn conditions, \textit{Original}, \textit{Neutral}, \textit{Retained},
and \textit{Revised}, single-turn controls isolating multi-turn
effects from task difficulty, and longer dialogues testing
whether interference accumulates or persists
(Section~\ref{sec:benchmark}).

\item \textbf{Empirical findings:} An analysis of accuracy degradation
across conditions and how it can deepen or persist as interaction
continues. Inactive-content error patterns are consistent with
\textit{mentioned-as-in-effect} confusion under \textit{Retained} and
\textit{Revised} (Section~\ref{sec:findings}).

\item \textbf{\method{}:} An on-policy self-distillation method that
selects the Teacher's single-turn task based on the user's decision
and trains the Student on the full dialogue, improving mean accuracy
over the corresponding base models by 10.81~pp across four models and four domains
(Section~\ref{sec:intent-opsd}).
\end{itemize}


%% file: content/3_benchmark.tex
\section{\benchmark{}: Evaluating Active Intent in Multi-Turn Dialogue}
\label{sec:benchmark}
This section describes \benchmark{} in detail. We first introduce the four multi-turn conditions (Section~\ref{sec:conditions}), then explain how these conditions are constructed from source tasks (Section~\ref{sec:construction}). Finally, we introduce single-turn controls to separate multi-turn effects from task difficulty and examine how conversational interference changes with interaction depth (Section~\ref{sec:single_controls}).

\subsection{Multi-Turn Conditions}
\label{sec:conditions}
Following \citet{laban2026lost}, we consider tasks whose requirements
are gradually revealed over multiple turns. We split each source task
into information shards and reveal them one at a time through user
messages $u_k$, preserving task information while changing only its
presentation. The first user message introduces the task goal, and later
messages provide the remaining details. After each user message $u_k$,
the assistant generates a response $a_k$ conditioned on the
user--assistant history
\begin{equation}
H_k = (u_1, a_1, \ldots, u_{k-1}, a_{k-1}, u_k).
\end{equation}
Beyond revealing task information, users may clarify earlier requirements or propose changes. We call the requirements that remain in effect \textit{\textbf{active intent}}. To study how models track them, we revise one requirement per source task, with reference answers $y^{(0)}$ and $y^{(1)}$ for the original and revised tasks. We then form four matched multi-turn conditions by continuing disclosure normally or inserting a two-user-message event after the selected requirement is introduced:
\begin{itemize}[leftmargin=2.5mm,labelsep=3pt,parsep=2pt]
\item \textbf{Original ($M_{\mathrm{Orig}}$):}
no event is inserted; disclosure continues normally.

\item \textbf{Neutral ($M_{\mathrm{Neu}}$):}
two clarifications are inserted; the requirement remains unchanged.

\item \textbf{Retained ($M_{\mathrm{Ret}}$):}
a change is proposed and rejected; the original requirement remains active.

\item \textbf{Revised ($M_{\mathrm{Rev}}$):}
the same proposed change is accepted; the revised requirement becomes active.
\end{itemize}

Figure~\ref{fig:construction} illustrates the protocol. In this example,
the selected requirement is introduced at $U_2$, followed by a
two-user-message event at $U_3$--$U_4$. \textit{Neutral} adds clarification
without proposing a change, while \textit{Retained} and \textit{Revised}
use the same proposal, followed by rejection or acceptance at
$U_4$. The dialogue then continues with the remaining task information.
Only the final response is scored. \textit{Original}, \textit{Neutral},
and \textit{Retained} are evaluated against the original task requirements
and reference answer $y^{(0)}$, while \textit{Revised} is evaluated
against the revised requirements and $y^{(1)}$.

\begin{figure}[t]
\centering

\includegraphics[width=1.0\linewidth]{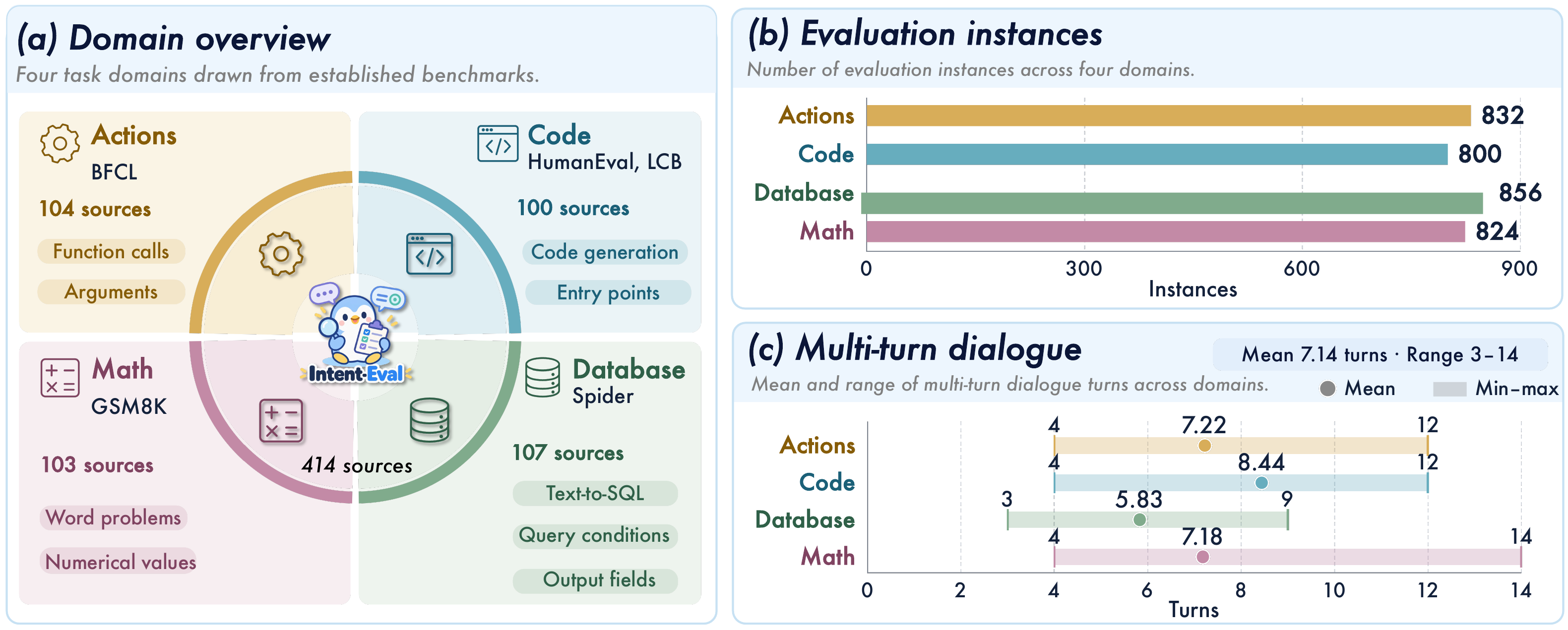}
\caption{Overview and statistics of \benchmark{}. The benchmark comprises 414 source tasks across four domains. Each source task has four multi-turn conditions and four single-turn controls, yielding 3,312 evaluation instances per model.}
\label{fig:dataset}
\end{figure}

\subsection{Data and Construction}
\label{sec:construction}

All 414 source tasks are adapted from the \textsc{LiC} benchmark~\citep{laban2026lost},
spanning four diverse domains: \domainlabel{Actions} for tool calling from
BFCL~\citep{patil2025bfcl}, \domainlabel{Code} for program synthesis from
HumanEval~\citep{chen2021evaluating} and LiveCodeBench~\citep{jain2025livecodebench},
\domainlabel{Database} for database querying from Spider~\citep{yu2018spider},
and \domainlabel{Math} for grade-school math from GSM8K~\citep{cobbe2021training}.
Figure~\ref{fig:dataset} summarizes the domain distributions, instance counts, and dialogue turn statistics. Each source yields four multi-turn conditions and four single-turn
controls, for 3{,}312 evaluated instances per model.
We construct the matched multi-turn conditions in four steps,
covering local revisions, reference answers, and condition-specific
dialogue events:

\noindent\textbf{\uppercase\expandafter{\romannumeral1}. Candidate enumeration.}
We apply domain-specific editing rules to the source tasks to generate
candidate revisions: changes to tool arguments in \textbf{\domainlabel{Actions}}, entry-point
names in \textbf{\domainlabel{Code}}, query conditions or output fields in \textbf{\domainlabel{Database}}, and
numerical values in \textbf{\domainlabel{Math}}.

\noindent\textbf{\uppercase\expandafter{\romannumeral2}. Constraint filtering.} 
Each candidate must satisfy three constraints: (1)~\textit{Unambiguity}: the change specifies exactly
which requirement is replaced and what replaces it;
(2)~\textit{Atomicity}: the change affects only one requirement
while preserving all other requirements; and
(3)~\textit{Verifiability}: the revised task has a reference answer
or output contract that can be checked by the task-specific verifier.

\noindent\textbf{\uppercase\expandafter{\romannumeral3}. Reference answer construction.}
After filtering, we select and apply one revision per source task,
then construct the corresponding reference answer $y^{(1)}$.

\noindent\textbf{\uppercase\expandafter{\romannumeral4}. Human review and rendering.}
We review revisions for clarity and consistency with other requirements,
then render approved revisions as shared proposals rejected in
\textit{Retained} or accepted in \textit{Revised}.
\textit{Neutral} adds two questions clarifying the same requirement
without proposing changes.

Appendix~\ref{app:construction} provides further construction details and a worked example.

\subsection{Controls and Interaction Extensions}
\label{sec:single_controls}

\paragraph{Single-turn controls.}

Our single-turn controls present complete task information
in one user message to
\ding{182}~separate revised-task difficulty from multi-turn effects; and
\ding{183}~examine whether accepted or rejected
proposals affect
performance without multi-turn interaction.
We first define two controls that state the task directly:
$S_{\mathrm{Orig}}$ presents the original task, while
$S_{\mathrm{Dir}}$ (Direct Revised) incorporates the revision
without a proposal or decision.
We then construct two decision-bearing controls by appending the same
proposal to $S_{\mathrm{Orig}}$, followed by acceptance or
rejection:
\begin{equation}
S_{\mathrm{Rev}} = S_{\mathrm{Orig}}\Vert
[\textsc{Propose}(r'),\textsc{Accept}],
\qquad
S_{\mathrm{Ret}} = S_{\mathrm{Orig}}\Vert
[\textsc{Propose}(r'),\textsc{Reject}],
\label{eq:single_controls}
\end{equation}
where $r'$ denotes the proposed replacement requirement and $\Vert$
denotes concatenation within the same user message.
In Figure~\ref{fig:construction}(a), $S_{\mathrm{Dir}}$ replaces
``\intentblue{4 bags}'' with ``\intentorange{2 bags}'' in the original prompt.
By contrast, $S_{\mathrm{Rev}}$ and $S_{\mathrm{Ret}}$ retain
``\intentblue{4 bags}'' and append the proposal to use
``\intentorange{2 bags}'', followed by \intentgreen{acceptance} or
\intentred{rejection}, respectively.
All other task information remains unchanged.
Thus, the original task remains in effect in $S_{\mathrm{Orig}}$
and $S_{\mathrm{Ret}}$, which use reference answer $y^{(0)}$;
the revised task is in effect in $S_{\mathrm{Dir}}$
and $S_{\mathrm{Rev}}$, which use $y^{(1)}$.
Accordingly, we match each multi-turn condition to a single-turn control with the same active requirements: $\{M_{\mathrm{Orig}},M_{\mathrm{Neu}},M_{\mathrm{Ret}}\}\!\boldsymbol{\to}\!S_{\mathrm{Orig}}$ and $M_{\mathrm{Rev}}\!\boldsymbol{\to}\!S_{\mathrm{Dir}}$.
We compare the single-turn controls in two ways:

\begin{itemize}[leftmargin=*,itemsep=0.5pt,parsep=0pt]
\item \textbf{Task difficulty ($S_{\mathrm{Orig}}$ vs.\ $S_{\mathrm{Dir}}$):}
Tests whether the revised task is harder when stated directly.

\item \textbf{Decision effects ($S_{\mathrm{Dir}}$ vs.\ $S_{\mathrm{Rev}}$;
$S_{\mathrm{Orig}}$ vs.\ $S_{\mathrm{Ret}}$):}
Tests whether expressing the same active task through a proposal
and decision affects performance relative to stating it directly.
\end{itemize}

\paragraph{Extended interaction.}
We use two protocols to test whether interference accumulates or persists
during interaction, with added-block count
$k$ from $0$ to $4$ (Appendix~\ref{app:interaction_protocols}
provides construction details):
\begin{itemize}[leftmargin=*,itemsep=0.5pt,parsep=0pt]
\item \textbf{Accumulation ($\text{Neutral}\times k$ vs.\ $\text{Retained}\times k$):}
Starting from \textit{Original} ($k=0$), we insert $k$ clarification
blocks or $k$ proposal-and-rejection blocks about the same requirement.
Each rejection block contains a distinct proposal.
Both paths preserve the original task and reference answer,
allowing us to compare how performance changes as clarification
or rejection accumulates.

\item \textbf{Persistence ($\{\text{Retained},\text{Revised}\}\to\text{Neutral}(k)$):}
We add $k$ neutral clarification blocks after either decision,
keeping the active task unchanged.
Comparison with each condition's $k=0$ baseline tests whether
interference persists during clarification and how it affects
performance.
\end{itemize}

%% file: content/4_method.tex
\section{\method{}: Learning Active Intent from Multi-Turn Dialogue}
\label{sec:intent-opsd}
\begin{figure}[t]
\centering

\includegraphics[width=1.0\linewidth]{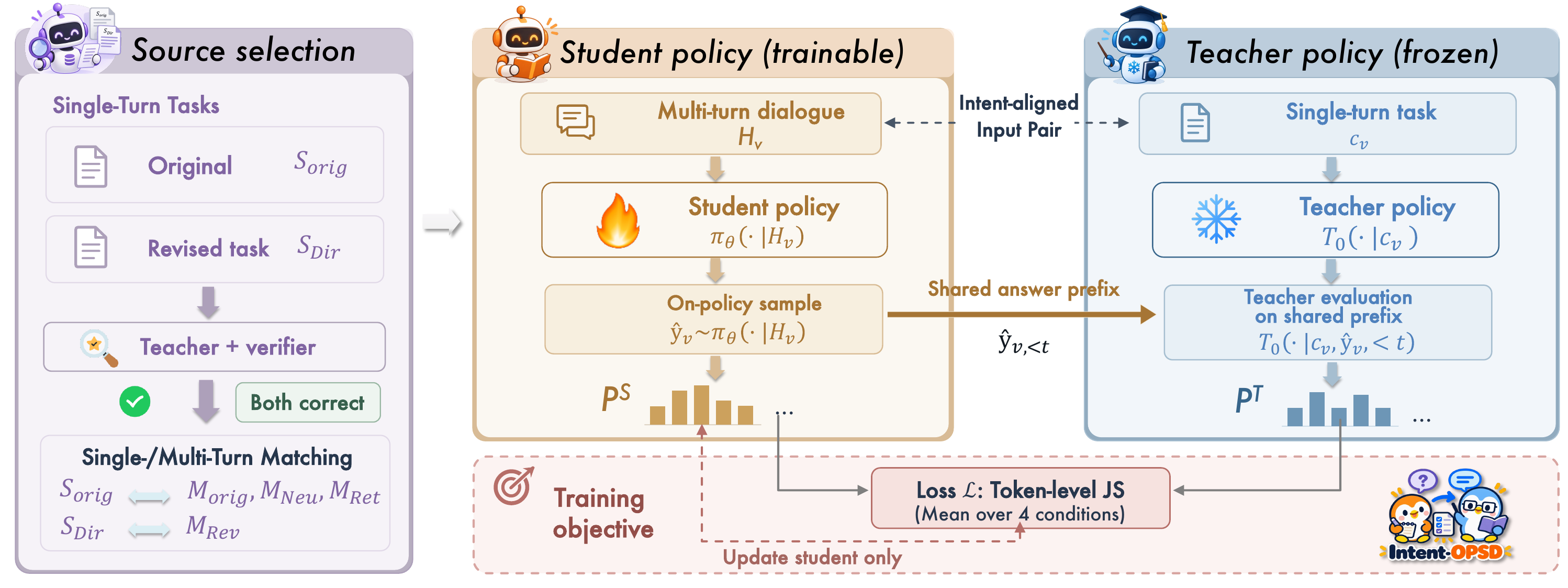}
\vspace{-6mm} 
\caption{\method{}: the frozen Teacher sees the active task; the Student sees the dialogue.}
\vspace{-2mm}
\label{fig:opsd_overview}
\end{figure}

Building on the active-task matching established in
Section~\ref{sec:single_controls}, \benchmark{} provides a complete
single-turn task for each multi-turn condition, specifying the
same active requirements.
As shown in Figure~\ref{fig:opsd_overview},
\method{} uses this single-turn task for supervision: the Student
receives the full dialogue, while a frozen Teacher receives the
corresponding complete task prompt.

\subsection{Active-Task Supervision and Source Selection}
\label{sec:method_insight}
Our goal is to supervise task execution according to requirements
that remain in effect, rather than all requirements mentioned in the
dialogue.
For each multi-turn condition indexed by $v$, we use its corresponding complete
single-turn task as the Teacher prompt:
\begin{equation}
c_v =
\begin{cases}
S_{\mathrm{Dir}}, & \text{if } v = \mathrm{Rev},\\
S_{\mathrm{Orig}}, & \text{otherwise}.
\end{cases}
\label{eq:teacher_view}
\end{equation}
This selection follows the active task: rejecting a proposal leaves the
Teacher on the original task, while accepting it switches the Teacher
to the revised task.
The Teacher thus provides supervision from a complete statement of the
active requirements, guiding the Student to identify and integrate those
requirements from the full dialogue despite redundant, rejected, or
superseded content.

For each candidate source, we first evaluate the Teacher on both complete single-turn tasks, $S_{\mathrm{Orig}}$ and $S_{\mathrm{Dir}}$. We retain a source for training only if both responses terminate normally and pass their respective verifiers. This check confirms that the Teacher can solve both the original and revised tasks from complete single-turn prompts. During training, the Teacher uses these prompts to supervise the Student under the matching multi-turn conditions. Reference answers and verifier results are used only for source selection and are never provided to either model.

\subsection{On-Policy Self Distillation of Intent}

For each admitted source and condition $v$, let $H_v$ denote the
multi-turn dialogue history under $M_v$ before the final response.
The Student receives $H_v$, while the frozen Teacher receives the
corresponding single-turn Teacher prompt $c_v$.
$M_{\mathrm{Ret}}$ and $M_{\mathrm{Rev}}$ share the dialogue prefix
through the Student's response to the same proposal, then branch at the
user's decision.
The Student then samples the final answer on-policy $\hat{y}_v \sim \pi_\theta(\cdot \mid H_v)$.

At each final-answer position $t$, both policies evaluate the next-token
distribution using the same Student-generated answer prefix
$\hat{y}_{v,<t}$:
\begin{equation}
\begin{aligned}
p^S_{v,t}
&= \pi_\theta(\cdot \mid H_v,\hat{y}_{v,<t}),
\qquad
p^T_{v,t}
= T_0(\cdot \mid c_v,\hat{y}_{v,<t}),
\qquad
\mathcal{L}_v= \frac{1}{|\hat{y}_v|}
\sum_{t=1}^{|\hat{y}_v|}
\operatorname{JS}\!\left(p^S_{v,t},p^T_{v,t}\right),
\end{aligned}
\label{eq:opsd_loss}
\end{equation}
where $T_0$ is the frozen base-model Teacher and $\operatorname{JS}$ is Jensen--Shannon divergence over the full vocabulary~\citep{lin1991divergence}. The Teacher scores Student-generated prefixes without generating a separate answer. Intermediate replies remain in $H_v$ but receive no loss.

We then average the four condition losses equally:
\begin{equation}
\mathcal{L}
= \frac{1}{4}
\left(
\mathcal{L}_{\mathrm{Orig}}
+ \mathcal{L}_{\mathrm{Neu}}
+ \mathcal{L}_{\mathrm{Ret}}
+ \mathcal{L}_{\mathrm{Rev}}
\right).
\end{equation}
Only the Student is updated, while the Teacher remains frozen.
The updated Student generates subsequent rollouts, maintaining
on-policy training.
At inference, the Student responds directly from the dialogue without
the Teacher or the complete-task prompt.

%% file: content/5_experiments.tex
\section{Experiments}
\label{sec:experiments}
\label{sec:findings}

In this section, we study accuracy, error patterns, continued interaction, and decision-conditioned distillation through four research questions:

\begin{itemize}[leftmargin=2.5mm,labelsep=3pt,parsep=4pt]
\item \textbf{RQ1:} How does multi-turn accuracy vary under different matching conditions? \textit{(\textbf{Phenomenon})}
\item \textbf{RQ2:} What characterizes errors under \textit{Retained} and \textit{Revised}? \textit{(\textbf{Diagnosis})}
\item \textbf{RQ3:} Does interference accumulate or persist as the dialogue continues? \textit{(\textbf{Depth})}
\item \textbf{RQ4:} Can decision-conditioned distillation improve multi-turn accuracy? \textit{(\textbf{Improvement})}
\end{itemize}

\input{tables/main_results}

\textbf{Experimental settings.}\label{sec:protocol}
We evaluate ten LLMs from six providers: \qwen~Qwen (Qwen3.6-27B, Qwen3-8B, Qwen3-4B-Instruct-2507, and Qwen2.5-7B-Instruct), \llama~Meta's Llama (Llama-3.1-8B), \openai~OpenAI (GPT-5.6-Luna), \deepseek~DeepSeek (DeepSeek-V4-Flash-0731), \claude~Anthropic (Claude-Sonnet-5), and \gemini~Google (Gemini-3.7-Flash and Gemma-4-26B-A4B). \textbf{RQ1} reports eight models in the main table, with earlier Qwen models in the appendix. Each model runs four single-turn controls and four multi-turn conditions. Multi-turn runs use a fixed Qwen3.6-27B user simulator, with assistants using dialogue history (see Appx.~\ref{app:simulator_reliability} for simulator reliability). \textbf{RQ2} pairs outputs by source and model across ten models, while \textbf{RQ3} varies interaction depth.

For \textbf{RQ4}, we compare \method{} with the Base and matched supervised fine-tuning (SFT). The main comparison uses Qwen3-8B, Gemma-4-26B-A4B, Qwen2.5-7B-Instruct, and Llama-3.1-8B. Each domain provides 500 candidate training sources and 50 validation sources, with no overlap with evaluation sources. We train on candidates for which the Teacher solves both task variants correctly. More detailed evaluation and training settings can be found in Appendices~\ref{app:evaluation_settings}, \ref{app:interaction_protocols}, and~\ref{app:opsd_training}.


\subsection{Multi-turn accuracy across conditions (RQ1)}
\label{sec:multi_results}

\paragraph{Overall performance.}
Table~\ref{tab:main_results} compares single-turn controls and multi-turn
conditions across eight models and four domains. We observe three patterns.

\textbf{(1) The multi-turn gap decomposes into disclosure and conversational overhead.}
As in prior work, distributing the same task across turns lowers mean
accuracy by 36.30~pp. More surprisingly, adding \textit{Neutral}
clarification causes a further 4.43~pp drop, even though it changes
neither the active task nor the reference answer. This pattern appears
across all eight models, suggesting that additional discussion can
interfere with task execution even when it introduces no new
requirement.

\textbf{(2) Decision-induced confusion.}
Introducing a proposed change lowers multi-turn accuracy by
5.55~pp when accepted and 8.06~pp when rejected, relative to
\textit{Original}. Compared with \textit{Neutral}, the additional drops are 1.12 and
3.63~pp, respectively.
This is not simply because the revised task is harder: models perform
similarly when both tasks are stated directly.
Even in a single turn, \textit{Revised} and \textit{Retained}
reduce accuracy by 5.74 and 7.52~pp, respectively, compared with
stating the same active task directly. Thus, models struggle to determine
what remains in effect after a decision.

\textbf{(3) Frontier models show that strong multi-turn robustness is achievable in Code.}
GPT-5.6 and DeepSeek-V4 retain at least 85\% accuracy across all multi-turn Code conditions,
contrasting with the larger degradation observed in Database, Actions, and Math.
This suggests that recent progress in conversational and agentic coding may improve multi-turn robustness. However, the advantage is highly
domain-specific: similarly strong robustness does not emerge in the other structured tasks.

\subsection{Characterizing errors under \textit{Retained} and \textit{Revised} (RQ2)}
\label{sec:single_results}

\begin{figure}[htbp!]
\centering
\vspace{-3mm}
\includegraphics[width=\linewidth]{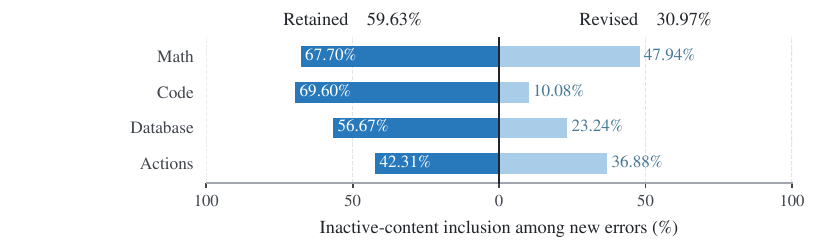}
\vspace{-5mm}
\caption{Inactive content in errors.
Bars show inclusion rates among new \textit{Retained} (left) and \textit{Revised} (right)
errors, pooled over ten models; headers aggregate all domains.}
\label{fig:inactive_content_inclusion}
\vspace{-2mm}
\end{figure}

\textbf{Settings.} To diagnose the accuracy gap, we analyze multi-turn cases
that succeed in \textit{Original} but fail in \textit{Retained} or
\textit{Revised}, checking for inactive content in their final responses. Detection details, per-model results for all ten models, and
illustrative examples appear in Appendix~\ref{app:error_analysis}.

\textbf{(1) Errors often include inactive content.}
Under \textit{Retained}, inactive content is the rejected proposal;
under \textit{Revised}, it is the original requirement replaced by the
proposal. As shown in Figure~\ref{fig:inactive_content_inclusion},
inactive content appears in 59.63\% of new \textit{Retained} errors
and 30.97\% of new \textit{Revised} errors across ten models.
These results show that inactive content often appears in these
erroneous responses, particularly after rejection, a pattern
consistent with \textit{mentioned-as-in-effect} confusion.

\textbf{(2) Math shows relatively high inactive-content inclusion under both decisions.}
In Math, these rates reach 67.70\% under \textit{Retained}
and 47.94\% under \textit{Revised}. This may reflect
the step-by-step nature of mathematical reasoning: changing
or restoring a numerical premise can require recomputing
several intermediate results. If earlier calculations are reused
without being updated, rejected or superseded values may
persist in subsequent steps.

\begin{figure}[!htbp]
\centering
\includegraphics[width=\textwidth]{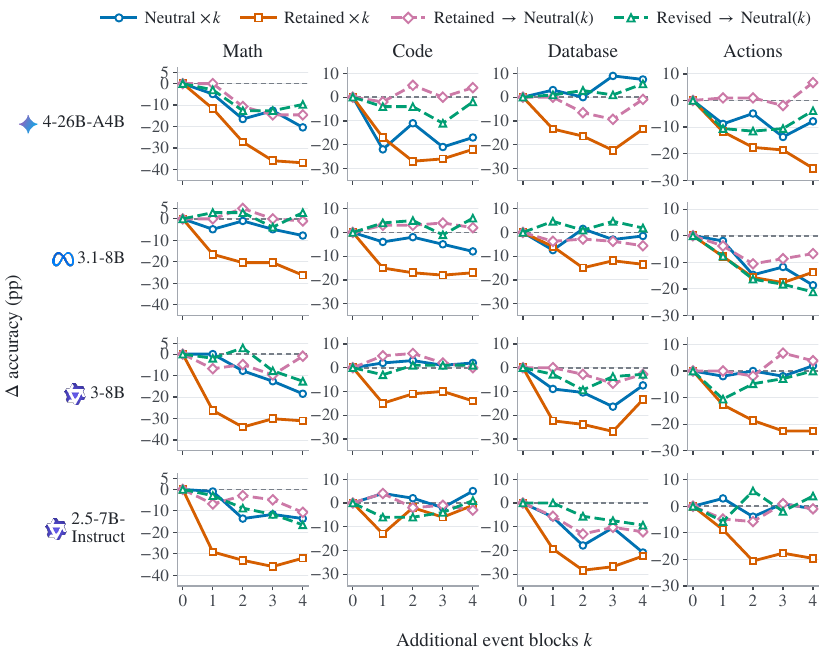}
\vspace{-8mm}
\caption{Accuracy change with depth relative to each curve's
$k=0$ baseline. Y-ranges are shared within domains.
Results for six additional models are in Appendix~\ref{app:interaction_depth_results}
(Figure~\ref{fig:supplementary_interaction_depth}).}
\vspace{-2mm}
\label{fig:all_depth_conditions}
\end{figure}

\subsection{Does interference accumulate or persist? (RQ3)}
\label{sec:suffix_results}
\label{sec:multik_results}

\paragraph{Settings.}
Accumulation uses the 372 of 414 sources that have four distinct proposals.
Added blocks leave the active requirements and reference answer unchanged.
Persistence uses 414 sources, adding \textit{Neutral} clarification after
\textit{Retained} or \textit{Revised} without changing the active requirements.

\textbf{(1) Accumulation: the first block causes the largest average drop, with further losses at greater depth.}
Across four models in Figure~\ref{fig:all_depth_conditions},
the first block produces the largest average single-step decline.
Accuracy generally continues to decline as blocks are added.
At the final depth, accuracy falls by 8.13~pp for \textit{Neutral} and
20.77~pp for \textit{Retained} relative to Original. As event
messages accumulate, repeatedly proposing and rejecting changes
produces greater degradation than clarification.

\textbf{(2) Persistence: continued clarification leaves accuracy below the decision baseline.}
Relative to each \textit{Retained} or \textit{Revised} baseline, mean accuracy
stays lower across all tested clarification depths. Adding four \textit{Neutral}
blocks lowers accuracy by 2.72~pp under \textit{Retained} and 3.50~pp under
\textit{Revised}. Further discussion can therefore hurt performance even after
the user has clearly accepted or rejected the proposal and the active task remains unchanged.

\subsection{Can decision-conditioned distillation improve accuracy? (RQ4)}
\label{sec:development_results}

\input{tables/method_results}

\textbf{(1) Active-task distillation improves multi-turn accuracy.}
\method{} improves mean accuracy over Base and SFT by 10.81 and
3.73~pp, respectively (Table~\ref{tab:method_results}).
It outperforms Base in all 16 model--domain pairs and SFT in 13.
Its mean gains over both baselines remain positive across tested depths.
Compared with SFT, gains are strong in \textit{Revised}, while mean \textit{Retained} accuracy is similar.
Gains over Base suggest the Teacher's active-task supervision helps
the Student follow active requirements, even when the dialogue
contains rejected or superseded content.

\textbf{(2) \method{} may help when intent tracking is a bottleneck.}
\method{} improves Actions accuracy by 21.46 pp over Base, where selecting the correct call arguments can fix the output. Math and Database may similarly benefit from tracking the numerical premises and query constraints that remain active. Code shows smaller gains, which may reflect its demands on program structure and logic. Even when the model correctly tracks the active requirement, structural or implementation errors can still cause the solution to fail. These results suggest that \method{} can be useful when models can solve the task but struggle to track which requirements to follow.

%% file: tables/main_results.tex
\begin{table}[b!]
\caption{
Model-by-domain accuracy (\%) across single-turn controls and the four multi-turn conditions. In Single, Direct Rev.\ states the revised task directly, while Revised includes the proposal and acceptance. Overall aggregates all 414 task sources across the eight models shown here; two additional models appear in Appendix Table~\ref{tab:supplementary_models}. Cell backgrounds indicate changes relative to Single Original: \protect\redsquare{} for declines, \protect\greensquare{} for gains, and \protect\yellowsquare{} for no change.
}
\vspace{-1mm}
\label{tab:main_results}
\centering
\scriptsize
\definecolor{resultLoss}{HTML}{e66565}
\definecolor{resultGain}{HTML}{90EE90}
\definecolor{resultTie}{HTML}{FFFFE0}
\setlength{\tabcolsep}{2.8pt}
\renewcommand{\arraystretch}{1.0}
\setlength{\aboverulesep}{0.3ex}
\setlength{\belowrulesep}{0.3ex}
\resizebox{1.0\linewidth}{!}{
\begin{tabular}{@{}llccccccc>{\columncolor{white}[\tabcolsep][0pt]}c@{}}
\toprule
\multirow{2}{*}{\textbf{Model}} & \multirow{2}{*}{\textbf{Domain}}
& \multicolumn{4}{c}{\textbf{Single}}
& \multicolumn{4}{c}{\textbf{Multi}} \\
\cmidrule(lr){3-6}\cmidrule(lr){7-10}
& & Original & Direct Rev. & Revised & Retained
& Original & Neutral & Revised & Retained \\
\midrule
\multirow{4}{*}{\qwen~3.6-27B}
& Math    & 98.06 & \cellcolor{resultLoss!10.84}97.09 & \cellcolor{resultLoss!19.26}87.38 & \cellcolor{resultLoss!14.21}93.20 & \cellcolor{resultLoss!31.04}73.79 & \cellcolor{resultLoss!32.73}71.84 & \cellcolor{resultLoss!36.93}66.99 & \cellcolor{resultLoss!41.98}61.17 \\
& Code    & 95.00 & \cellcolor{resultLoss!11.79}93.00 & \cellcolor{resultLoss!14.47}90.00 & \cellcolor{resultLoss!16.26}88.00 & \cellcolor{resultLoss!22.53}81.00 & \cellcolor{resultLoss!27.00}76.00 & \cellcolor{resultLoss!23.42}80.00 & \cellcolor{resultLoss!30.58}72.00 \\
& DB      & 92.52 & \cellcolor{resultLoss!10.85}91.59 & \cellcolor{resultLoss!10.85}91.59 & \cellcolor{resultLoss!14.29}87.85 & \cellcolor{resultLoss!55.50}42.99 & \cellcolor{resultLoss!54.64}43.93 & \cellcolor{resultLoss!56.36}42.06 & \cellcolor{resultLoss!59.79}38.32 \\
& Actions & 96.15 & \cellcolor{resultLoss!10.85}95.19 & \cellcolor{resultTie!35}96.15 & \cellcolor{resultLoss!19.34}85.58 & \cellcolor{resultLoss!51.65}49.04 & \cellcolor{resultLoss!55.90}44.23 & \cellcolor{resultLoss!53.34}47.12 & \cellcolor{resultLoss!52.50}48.08 \\
\midrule
\multirow{4}{*}{\qwen~3-8B}
& Math    & 92.23 & \cellcolor{resultLoss!11.79}90.29 & \cellcolor{resultLoss!25.21}75.73 & \cellcolor{resultLoss!32.37}67.96 & \cellcolor{resultLoss!35.95}64.08 & \cellcolor{resultLoss!35.95}64.08 & \cellcolor{resultLoss!44.89}54.37 & \cellcolor{resultLoss!60.11}37.86 \\
& Code    & 73.00 & \cellcolor{resultGain!12.33}75.00 & \cellcolor{resultLoss!46.10}42.00 & \cellcolor{resultLoss!12.33}71.00 & \cellcolor{resultLoss!47.26}41.00 & \cellcolor{resultLoss!44.93}43.00 & \cellcolor{resultLoss!44.93}43.00 & \cellcolor{resultLoss!64.73}26.00 \\
& DB      & 85.98 & \cellcolor{resultLoss!10.92}85.05 & \cellcolor{resultLoss!21.08}74.77 & \cellcolor{resultLoss!45.11}50.47 & \cellcolor{resultLoss!57.12}38.32 & \cellcolor{resultLoss!69.13}26.17 & \cellcolor{resultLoss!62.66}32.71 & \cellcolor{resultLoss!75.00}17.76 \\
& Actions & 97.12 & \cellcolor{resultLoss!10.85}96.15 & \cellcolor{resultLoss!21.79}83.65 & \cellcolor{resultLoss!20.10}85.58 & \cellcolor{resultLoss!49.55}51.92 & \cellcolor{resultLoss!51.24}50.00 & \cellcolor{resultLoss!54.60}46.15 & \cellcolor{resultLoss!60.50}39.42 \\
\midrule
\multirow{4}{*}{\llama~3.1-8B}
& Math    & 75.73 & \cellcolor{resultLoss!13.27}72.82 & \cellcolor{resultLoss!25.25}62.14 & \cellcolor{resultLoss!23.08}64.08 & \cellcolor{resultLoss!56.86}33.98 & \cellcolor{resultLoss!62.31}29.13 & \cellcolor{resultLoss!69.94}22.33 & \cellcolor{resultLoss!75.00}17.48 \\
& Code    & 45.00 & \cellcolor{resultGain!13.78}47.00 & \cellcolor{resultLoss!13.78}43.00 & \cellcolor{resultLoss!49.67}24.00 & \cellcolor{resultLoss!47.78}25.00 & \cellcolor{resultLoss!55.33}21.00 & \cellcolor{resultLoss!64.78}16.00 & \cellcolor{resultLoss!75.00}10.00 \\
& DB      & 79.44 & \cellcolor{resultLoss!11.01}78.50 & \cellcolor{resultLoss!13.00}76.64 & \cellcolor{resultLoss!48.00}43.93 & \cellcolor{resultLoss!71.00}22.43 & \cellcolor{resultLoss!69.00}24.30 & \cellcolor{resultLoss!75.00}16.82 & \cellcolor{resultLoss!74.00}19.63 \\
& Actions & 93.27 & \cellcolor{resultTie!35}93.27 & \cellcolor{resultLoss!29.27}72.12 & \cellcolor{resultLoss!21.39}80.77 & \cellcolor{resultLoss!44.18}55.77 & \cellcolor{resultLoss!46.80}52.88 & \cellcolor{resultLoss!51.19}48.08 & \cellcolor{resultLoss!51.19}48.08 \\
\midrule
\multirow{4}{*}{\gemini~4-26B-A4B}
& Math    & 97.09 & \cellcolor{resultGain!10.85}98.06 & \cellcolor{resultLoss!18.50}87.38 & \cellcolor{resultLoss!14.25}92.23 & \cellcolor{resultLoss!33.80}69.90 & \cellcolor{resultLoss!38.05}65.05 & \cellcolor{resultLoss!37.20}66.02 & \cellcolor{resultLoss!44.00}58.25 \\
& Code    & 92.00 & \cellcolor{resultLoss!10.92}91.00 & \cellcolor{resultLoss!29.40}71.00 & \cellcolor{resultLoss!12.77}89.00 & \cellcolor{resultLoss!24.78}76.00 & \cellcolor{resultLoss!45.11}54.00 & \cellcolor{resultLoss!32.17}68.00 & \cellcolor{resultLoss!40.49}59.00 \\
& DB      & 93.46 & \cellcolor{resultLoss!12.56}90.65 & \cellcolor{resultLoss!14.25}88.79 & \cellcolor{resultLoss!11.70}91.59 & \cellcolor{resultLoss!56.75}42.06 & \cellcolor{resultLoss!54.20}44.86 & \cellcolor{resultLoss!56.75}42.06 & \cellcolor{resultLoss!62.70}35.51 \\
& Actions & 98.08 & \cellcolor{resultTie!35}98.08 & \cellcolor{resultLoss!15.00}92.31 & \cellcolor{resultLoss!15.00}92.31 & \cellcolor{resultLoss!60.00}40.38 & \cellcolor{resultLoss!67.50}31.73 & \cellcolor{resultLoss!63.33}36.54 & \cellcolor{resultLoss!70.00}28.85 \\
\midrule
\multirow{4}{*}{\openai~5.6-Luna}
& Math    & 95.15 & \cellcolor{resultGain!10.87}96.12 & \cellcolor{resultLoss!14.34}90.29 & \cellcolor{resultLoss!19.54}84.47 & \cellcolor{resultLoss!26.48}76.70 & \cellcolor{resultLoss!29.08}73.79 & \cellcolor{resultLoss!37.76}64.08 & \cellcolor{resultLoss!39.49}62.14 \\
& Code    & 96.00 & \cellcolor{resultLoss!13.54}92.00 & \cellcolor{resultLoss!12.66}93.00 & \cellcolor{resultLoss!21.51}83.00 & \cellcolor{resultLoss!12.66}93.00 & \cellcolor{resultLoss!13.54}92.00 & \cellcolor{resultLoss!12.66}93.00 & \cellcolor{resultLoss!19.74}85.00 \\
& DB      & 94.39 & \cellcolor{resultLoss!10.84}93.46 & \cellcolor{resultLoss!10.84}93.46 & \cellcolor{resultLoss!13.37}90.65 & \cellcolor{resultLoss!56.29}42.99 & \cellcolor{resultLoss!59.65}39.25 & \cellcolor{resultLoss!51.23}48.60 & \cellcolor{resultLoss!62.18}36.45 \\
& Actions & 86.54 & \cellcolor{resultTie!35}86.54 & \cellcolor{resultGain!14.72}91.35 & \cellcolor{resultGain!11.89}88.46 & \cellcolor{resultLoss!54.39}41.35 & \cellcolor{resultLoss!55.34}40.38 & \cellcolor{resultLoss!61.00}34.62 & \cellcolor{resultLoss!61.95}33.65 \\
\midrule
\multirow{4}{*}{\gemini~3.7-Flash}
& Math    & 97.09 & \cellcolor{resultGain!11.70}99.03 & \cellcolor{resultLoss!14.25}92.23 & \cellcolor{resultLoss!11.70}95.15 & \cellcolor{resultLoss!31.25}72.82 & \cellcolor{resultLoss!36.35}66.99 & \cellcolor{resultLoss!41.45}61.17 & \cellcolor{resultLoss!35.50}67.96 \\
& Code    & 99.00 & \cellcolor{resultLoss!13.43}95.00 & \cellcolor{resultLoss!23.74}83.00 & \cellcolor{resultTie!35}99.00 & \cellcolor{resultTie!35}99.00 & \cellcolor{resultLoss!36.62}68.00 & \cellcolor{resultLoss!57.22}44.00 & \cellcolor{resultLoss!14.29}94.00 \\
& DB      & 99.07 & \cellcolor{resultTie!35}99.07 & \cellcolor{resultLoss!13.21}95.33 & \cellcolor{resultLoss!10.81}98.13 & \cellcolor{resultLoss!58.91}42.06 & \cellcolor{resultLoss!57.31}43.93 & \cellcolor{resultLoss!62.12}38.32 & \cellcolor{resultLoss!63.73}36.45 \\
& Actions & 98.08 & \cellcolor{resultTie!35}98.08 & \cellcolor{resultGain!10.83}99.04 & \cellcolor{resultTie!35}98.08 & \cellcolor{resultLoss!62.50}37.50 & \cellcolor{resultLoss!66.67}32.69 & \cellcolor{resultLoss!60.84}39.42 & \cellcolor{resultLoss!59.16}41.35 \\
\midrule
\multirow{4}{*}{\deepseek~V4-Flash-0731}
& Math    & 98.06 & \cellcolor{resultLoss!10.84}97.09 & \cellcolor{resultLoss!13.37}94.17 & \cellcolor{resultLoss!14.21}93.20 & \cellcolor{resultLoss!46.19}56.31 & \cellcolor{resultLoss!45.35}57.28 & \cellcolor{resultLoss!47.87}54.37 & \cellcolor{resultLoss!47.87}54.37 \\
& Code    & 98.00 & \cellcolor{resultGain!10.87}99.00 & \cellcolor{resultGain!11.73}100.00 & \cellcolor{resultLoss!10.87}97.00 & \cellcolor{resultLoss!12.60}95.00 & \cellcolor{resultLoss!16.94}90.00 & \cellcolor{resultLoss!20.41}86.00 & \cellcolor{resultLoss!14.34}93.00 \\
& DB      & 92.52 & \cellcolor{resultGain!10.86}93.46 & \cellcolor{resultLoss!15.14}86.92 & \cellcolor{resultTie!35}92.52 & \cellcolor{resultLoss!61.51}36.45 & \cellcolor{resultLoss!57.22}41.12 & \cellcolor{resultLoss!65.80}31.78 & \cellcolor{resultLoss!62.38}35.51 \\
& Actions & 94.23 & \cellcolor{resultLoss!10.87}93.27 & \cellcolor{resultTie!35}94.23 & \cellcolor{resultTie!35}94.23 & \cellcolor{resultLoss!61.17}37.50 & \cellcolor{resultLoss!63.77}34.62 & \cellcolor{resultLoss!67.24}30.77 & \cellcolor{resultLoss!64.65}33.65 \\
\midrule
\multirow{4}{*}{\claude~Sonnet-5}
& Math    & 95.15 & \cellcolor{resultLoss!12.61}92.23 & \cellcolor{resultLoss!16.94}87.38 & \cellcolor{resultLoss!15.21}89.32 & \cellcolor{resultLoss!36.89}65.05 & \cellcolor{resultLoss!34.29}67.96 & \cellcolor{resultLoss!33.42}68.93 & \cellcolor{resultLoss!36.89}65.05 \\
& Code    & 97.00 & \cellcolor{resultLoss!14.38}92.00 & \cellcolor{resultTie!35}97.00 & \cellcolor{resultLoss!10.88}96.00 & \cellcolor{resultLoss!17.89}88.00 & \cellcolor{resultLoss!44.18}58.00 & \cellcolor{resultLoss!17.89}88.00 & \cellcolor{resultLoss!24.02}81.00 \\
& DB      & 94.39 & \cellcolor{resultLoss!10.84}93.46 & \cellcolor{resultLoss!10.84}93.46 & \cellcolor{resultLoss!11.68}92.52 & \cellcolor{resultLoss!57.12}42.06 & \cellcolor{resultLoss!53.77}45.79 & \cellcolor{resultLoss!57.97}41.12 & \cellcolor{resultLoss!63.86}34.58 \\
& Actions & 95.19 & \cellcolor{resultLoss!11.71}93.27 & \cellcolor{resultTie!35}95.19 & \cellcolor{resultTie!35}95.19 & \cellcolor{resultLoss!49.50}50.96 & \cellcolor{resultLoss!54.65}45.19 & \cellcolor{resultLoss!48.64}51.92 & \cellcolor{resultLoss!46.91}53.85 \\
\midrule
\multicolumn{2}{l}{Eight-model overall}
& 91.73 & \cellcolor{resultLoss!10.82}90.85 & \cellcolor{resultLoss!16.13}85.11 & \cellcolor{resultLoss!16.97}84.21 & \cellcolor{resultLoss!43.63}55.43 & \cellcolor{resultLoss!47.74}51.00 & \cellcolor{resultLoss!48.78}49.88 & \cellcolor{resultLoss!51.10}47.37 \\
\bottomrule
\end{tabular}
}

\end{table}

%% file: tables/method_results.tex
\begin{table}[!htbp]
\caption{Accuracy (\%) and \method{} depth gains (pp).
For accuracy, conditions and model--domain pairs have equal weight.
Colors show changes from Base (\protect\redsquare{} lower, \protect\greensquare{} higher, \protect\yellowsquare{} unchanged); shading reflects magnitude.
Gains use overall accuracy across the four interaction paths;
$k$ counts added event blocks.
Appendix~\ref{app:opsd_depth_results} provides detailed results by path and depth.}
\label{tab:method_results}
\centering
\footnotesize
\definecolor{methodGain}{HTML}{90EE90}
\definecolor{methodTie}{HTML}{FFFFE0}
\definecolor{methodLoss}{HTML}{e66565}
\setlength{\tabcolsep}{2.5pt}
\renewcommand{\arraystretch}{1}
\setlength{\aboverulesep}{0.3ex}
\setlength{\belowrulesep}{0.3ex}
\resizebox{1.0\linewidth}{!}{
\begin{tabular}{@{}llcccc@{\hspace{12pt}}cccc@{\hspace{12pt}}cccc@{}}
\toprule
\multirow{2}{*}{Model} & \multirow{2}{*}{Domain}
& \multicolumn{4}{c}{Base} & \multicolumn{4}{c}{SFT} & \multicolumn{4}{c}{\method{}} \\
\cmidrule(lr){3-6}\cmidrule(lr){7-10}\cmidrule(lr){11-14}
& & Orig. & Neut. & Revised & Retained
& Orig. & Neut. & Revised & Retained
& Orig. & Neut. & Revised & Retained \\
\midrule
\multirow{4}{*}{\qwen~3-8B} & Math & 64.08 & 64.08 & 54.37 & 37.86 & \cellcolor{methodLoss!13.86}61.17 & \cellcolor{methodLoss!13.86}61.17 & \cellcolor{methodGain!17.59}59.22 & \cellcolor{methodLoss!31.79}28.16 & \cellcolor{methodGain!12.58}66.02 & \cellcolor{methodGain!15.15}67.96 & \cellcolor{methodGain!25.18}64.08 & \cellcolor{methodLoss!42.69}23.30 \\
 & Code & 41.00 & 43.00 & 43.00 & 26.00 & \cellcolor{methodLoss!22.44}35.00 & \cellcolor{methodLoss!23.84}36.00 & \cellcolor{methodLoss!21.86}37.00 & \cellcolor{methodLoss!16.54}24.00 & \cellcolor{methodLoss!14.15}39.00 & \cellcolor{methodLoss!11.98}42.00 & \cellcolor{methodGain!13.95}45.00 & \cellcolor{methodGain!16.54}28.00 \\
 & DB & 38.32 & 26.17 & 32.71 & 17.76 & \cellcolor{methodGain!22.44}43.93 & \cellcolor{methodGain!49.46}38.32 & \cellcolor{methodGain!39.14}43.93 & \cellcolor{methodGain!23.42}20.56 & \cellcolor{methodGain!26.59}45.79 & \cellcolor{methodGain!70.71}44.86 & \cellcolor{methodGain!36.71}42.99 & \cellcolor{methodGain!54.74}27.10 \\
 & Actions & 51.92 & 50.00 & 46.15 & 39.42 & \cellcolor{methodGain!36.76}68.27 & \cellcolor{methodGain!45.96}71.15 & \cellcolor{methodLoss!36.56}31.73 & \cellcolor{methodGain!49.39}57.69 & \cellcolor{methodGain!33.61}66.35 & \cellcolor{methodGain!27.98}60.58 & \cellcolor{methodGain!47.19}66.35 & \cellcolor{methodGain!45.24}55.77 \\
\midrule
\multirow{4}{*}{\shortstack[l]{\gemini~4-26B-\\A4B}} & Math & 69.90 & 65.05 & 66.02 & 58.25 & \cellcolor{methodGain!11.18}70.87 & \cellcolor{methodGain!18.88}71.84 & \cellcolor{methodTie!35}66.02 & \cellcolor{methodGain!15.67}62.14 & \cellcolor{methodGain!24.17}81.55 & \cellcolor{methodGain!26.49}77.67 & \cellcolor{methodGain!18.75}72.82 & \cellcolor{methodGain!12.83}60.19 \\
 & Code & 76.00 & 54.00 & 68.00 & 59.00 & \cellcolor{methodLoss!13.36}73.00 & \cellcolor{methodGain!32.04}68.00 & \cellcolor{methodLoss!11.25}67.00 & \cellcolor{methodLoss!12.88}57.00 & \cellcolor{methodLoss!12.24}74.00 & \cellcolor{methodGain!30.46}67.00 & \cellcolor{methodLoss!11.25}67.00 & \cellcolor{methodLoss!12.88}57.00 \\
 & DB & 42.06 & 44.86 & 42.06 & 35.51 & \cellcolor{methodGain!49.67}61.68 & \cellcolor{methodGain!27.71}54.21 & \cellcolor{methodGain!19.44}46.73 & \cellcolor{methodGain!27.89}42.99 & \cellcolor{methodGain!55.33}64.49 & \cellcolor{methodGain!40.10}60.75 & \cellcolor{methodGain!27.00}50.47 & \cellcolor{methodGain!34.61}45.79 \\
 & Actions & 40.38 & 31.73 & 36.54 & 28.85 & \cellcolor{methodGain!75.00}84.62 & \cellcolor{methodGain!75.00}79.81 & \cellcolor{methodGain!75.00}65.38 & \cellcolor{methodGain!75.00}81.73 & \cellcolor{methodGain!75.00}82.69 & \cellcolor{methodGain!75.00}75.96 & \cellcolor{methodGain!75.00}83.65 & \cellcolor{methodGain!75.00}75.96 \\
\midrule
\multirow{4}{*}{\shortstack[l]{\qwen~2.5-7B-\\Instruct}} & Math & 49.51 & 48.54 & 46.60 & 20.39 & \cellcolor{methodLoss!11.67}48.54 & \cellcolor{methodLoss!25.30}39.81 & \cellcolor{methodLoss!24.17}38.83 & \cellcolor{methodTie!35}20.39 & \cellcolor{methodGain!31.67}62.14 & \cellcolor{methodGain!16.80}52.43 & \cellcolor{methodGain!11.77}47.57 & \cellcolor{methodGain!42.38}28.16 \\
 & Code & 40.00 & 44.00 & 34.00 & 27.00 & \cellcolor{methodGain!16.38}43.00 & \cellcolor{methodGain!13.86}46.00 & \cellcolor{methodGain!20.00}38.00 & \cellcolor{methodGain!38.33}36.00 & \cellcolor{methodGain!18.50}44.00 & \cellcolor{methodGain!21.59}50.00 & \cellcolor{methodLoss!15.00}32.00 & \cellcolor{methodLoss!16.30}25.00 \\
 & DB & 40.19 & 34.58 & 33.64 & 19.63 & \cellcolor{methodGain!13.95}42.06 & \cellcolor{methodLoss!30.68}26.17 & \cellcolor{methodGain!28.89}41.12 & \cellcolor{methodLoss!58.57}8.41 & \cellcolor{methodGain!31.74}50.47 & \cellcolor{methodLoss!14.59}32.71 & \cellcolor{methodGain!45.42}47.66 & \cellcolor{methodGain!14.05}20.56 \\
 & Actions & 47.12 & 50.00 & 49.04 & 38.46 & \cellcolor{methodGain!42.96}65.38 & \cellcolor{methodGain!36.15}65.38 & \cellcolor{methodLoss!35.00}34.62 & \cellcolor{methodGain!39.75}51.92 & \cellcolor{methodGain!42.96}65.38 & \cellcolor{methodGain!34.52}64.42 & \cellcolor{methodLoss!13.33}47.12 & \cellcolor{methodLoss!24.88}31.73 \\
\midrule
\multirow{4}{*}{\llama~3.1-8B} & Math & 33.98 & 29.13 & 22.33 & 17.48 & \cellcolor{methodGain!51.29}50.49 & \cellcolor{methodGain!38.33}38.83 & \cellcolor{methodLoss!17.39}20.39 & \cellcolor{methodGain!75.00}33.98 & \cellcolor{methodGain!61.00}54.37 & \cellcolor{methodGain!66.67}48.54 & \cellcolor{methodGain!54.35}33.98 & \cellcolor{methodGain!66.67}29.13 \\
 & Code & 25.00 & 21.00 & 16.00 & 10.00 & \cellcolor{methodLoss!16.80}23.00 & \cellcolor{methodGain!18.10}23.00 & \cellcolor{methodLoss!47.19}9.00 & \cellcolor{methodGain!75.00}18.00 & \cellcolor{methodGain!16.80}27.00 & \cellcolor{methodGain!22.14}24.00 & \cellcolor{methodGain!41.88}22.00 & \cellcolor{methodGain!61.00}16.00 \\
 & DB & 22.43 & 24.30 & 16.82 & 19.63 & \cellcolor{methodGain!75.00}41.12 & \cellcolor{methodGain!42.69}33.64 & \cellcolor{methodGain!71.39}28.97 & \cellcolor{methodLoss!14.05}18.69 & \cellcolor{methodGain!75.00}47.66 & \cellcolor{methodGain!65.58}40.19 & \cellcolor{methodGain!75.00}38.32 & \cellcolor{methodGain!34.29}25.23 \\
 & Actions & 55.77 & 52.88 & 48.08 & 48.08 & \cellcolor{methodGain!42.24}76.92 & \cellcolor{methodGain!47.09}75.96 & \cellcolor{methodGain!32.10}60.58 & \cellcolor{methodGain!38.90}64.42 & \cellcolor{methodGain!46.64}79.81 & \cellcolor{methodGain!53.27}79.81 & \cellcolor{methodGain!20.20}53.85 & \cellcolor{methodGain!45.70}68.27 \\
\midrule
\multicolumn{2}{l}{Mean} & 46.10 & 42.71 & 40.96 & 31.46 & \cellcolor{methodGain!27.44}55.57 & \cellcolor{methodGain!28.16}51.83 & \cellcolor{methodGain!14.30}43.03 & \cellcolor{methodGain!30.73}39.13 & \cellcolor{methodGain!34.55}59.42 & \cellcolor{methodGain!35.57}55.55 & \cellcolor{methodGain!30.69}50.93 & \cellcolor{methodGain!29.23}38.57 \\
\addlinespace[2pt]
\multicolumn{2}{l}{\textbf{Overall Avg.}} & \multicolumn{4}{c@{\hspace{12pt}}}{\textbf{40.31}} & \multicolumn{4}{c@{\hspace{12pt}}}{\cellcolor{methodGain!24.94}\textbf{47.39}} & \multicolumn{4}{c}{\cellcolor{methodGain!32.80}\textbf{51.12}} \\
\midrule
\multicolumn{2}{l}{\textbf{Depth gain (pp)}} & \multicolumn{3}{c}{$k=1$} & \multicolumn{3}{c}{$k=2$} & \multicolumn{3}{c}{$k=3$} & \multicolumn{3}{c}{$k=4$} \\
\addlinespace[2pt]
\multicolumn{2}{l}{vs. Base} & \multicolumn{3}{c}{\cellcolor{methodGain!32.71}+9.61} & \multicolumn{3}{c}{\cellcolor{methodGain!39.72}+11.74} & \multicolumn{3}{c}{\cellcolor{methodGain!40.79}+11.67} & \multicolumn{3}{c}{\cellcolor{methodGain!37.34}+10.64} \\
\multicolumn{2}{l}{vs. SFT} & \multicolumn{3}{c}{\cellcolor{methodGain!15.80}+2.91} & \multicolumn{3}{c}{\cellcolor{methodGain!17.93}+3.86} & \multicolumn{3}{c}{\cellcolor{methodGain!15.12}+2.49} & \multicolumn{3}{c}{\cellcolor{methodGain!14.42}+2.16} \\
\bottomrule
\end{tabular}
}
\end{table}

%% file: content/2_relatedWork.tex
\section{Related Work}
\label{sec:related_work}


\textbf{Multi-turn Dialogue.}
Early multi-turn benchmarks primarily evaluate instruction retention, contextual reasoning, and response refinement~\citep{bai2024mtbench101,kwan2024mteval,deshpande2025multichallenge}. LiC further reports performance degradation as tasks unfold across multiple turns~\citep{laban2026lost}. Moving beyond general multi-turn competence, \textit{Intent Mismatch} , which studies intent alignment by reconstructing instructions from dialogue context~\citep{liu2026intentmismatch}, is a more challenging and fundamental problem. Recent benchmarks examine evolving user requirements. SEQUOR~\citep{canaverde2026sequor} and Evolving User Intent~\citep{tack2026evolving} study requirement changes, SpecPath~\citep{wu2026specpath} compares interaction histories with equivalent final specifications that may exclude canceled requirements, and Trip+~\citep{chen2026tripplus} evaluates adaptation to changing travel needs while preserving valid commitments. \textit{Unlike prior works that focus on particular forms or domains of evolving intent, our Intent-Eval systematically studies how conversational decisions alter the effective task state, jointly examining clarification, rejection, and revision with matched single-turn controls and extended dialogues.}


\textbf{On-Policy Distillation.}
On-policy distillation supervises Student-generated prefixes with Teacher distributions, enabling knowledge transfer directly on the Student's trajectory~\citep{agarwal2024gkd}. Building on this paradigm, Self-Distilled Reasoner introduces on-policy self-distillation~\citep{zhao2026selfdistilled}, where a same-model Teacher is provided with reference solutions. On-policy context distillation~\citep{ye2026opcd} further studies asymmetric context, providing the Teacher with information unavailable to the Student. More closely related to multi-turn dialogue, Found in Conversation~\citep{chen2026found} transfers single-turn competence to conversational settings, while MAIGO~\citep{zheng2026maigo} performs self-distillation using supervision derived from cleaned dialogue histories. \textit{In contrast, \method{} systematically conditions Teacher supervision on the user's conversational decision, constructing the corresponding single-turn task from the requirements that remain in effect and distilling this supervision into a Student operating on the full dialogue.}

%% file: content/6_conclusion.tex

\section{Conclusion}
\label{sec:conclusion}
In this work, we systematically study how frontier models handle evolving user intent in multi-turn interactions. We introduce \benchmark{}, which compares clarification, rejection, and revision with matched single-turn controls across four domains, revealing performance degradation that can deepen or persist over longer dialogues. Our analysis further identifies \textit{mentioned-as-in-effect} confusion, where models incorrectly treat inactive content as still operative. To address this failure, we propose \method{}, which aligns the Teacher's single-turn task with the requirements that remain in effect while training the Student on the full dialogue. \method{} significantly mitigates this issue without requiring the teacher model during inference. Together, our findings highlight the importance of distinguishing what was \textit{said} from what remains \textit{in effect}. Future work can extend this study to longer and more open-ended interactions with richer forms of intent evolution.

%% file: appendix/0_appendix_list.tex



\onecolumn
\renewcommand{\cftsecfont}{\bfseries}
\begin{center}
    \Large{\sc\Huge Supplementary material\\\large ~\\You Changed Your Mind, The Model Didn’t: \\Demystifying Intent in Multi-Turn Dialogue}\\
\end{center}
\vskip 6mm
\startcontents[sections]\vbox{\sc\Large Table of Contents}
\vspace{5mm}
\hrule height .8pt
\vspace{-2mm}
\printcontents[sections]{l}{1}{\setcounter{tocdepth}{2}}
\vspace{4mm}
\hrule height .8pt
\vskip 10mm

\clearpage
\input{appendix/a_construction}
\FloatBarrier
\input{appendix/c_evaluation}
\FloatBarrier
\input{appendix/b_conditions}
\FloatBarrier
\input{appendix/d_training}
\FloatBarrier
\input{appendix/e_supplementary_results}
\FloatBarrier
\input{appendix/f_discussion}
\FloatBarrier

%% file: appendix/a_construction.tex
\section{Construction Details, Scoring, and Examples}
\label{app:construction}

Section~\ref{app:construction_scoring} supplements the construction procedure
in Section~\ref{sec:construction} (Steps~I--IV) with domain-specific editing
rules, candidate checks and selection, reference construction, and condition
rendering. It also describes scoring for each domain.
Section~\ref{app:worked_example} follows one task through the complete
process, including the messages and reference answers used in evaluation.

\subsection{Construction details}
\label{app:construction_scoring}

Table~\ref{tab:construction_rules} summarizes the edits and reference updates.
We apply the construction constraints in Section~\ref{sec:construction}.
We discard candidates with invalid references or whose original and revised
task states cannot be distinguished by the applicable verifier or
output-contract checks.

\begin{table}[htbp]
\centering
\small
\setlength{\tabcolsep}{6pt}
\renewcommand{\arraystretch}{1.15}
\caption{Domain-specific revisions and reference updates.}
\label{tab:construction_rules}
\begin{tabularx}{\linewidth}{@{}l*{2}{>{\raggedright\arraybackslash}X}@{}}
\toprule
Domain & Revised requirement & Reference update \\
\midrule
\domainlabel{Actions} & Tool argument in specified calls & Replace the affected argument values \\
\addlinespace[5pt]
\domainlabel{Code} & Entry-point function or method name & Rename the callable in the contract and available reference code \\
\addlinespace[5pt]
\domainlabel{Database} & Query condition or output field & Update the corresponding SQL clause \\
\addlinespace[5pt]
\domainlabel{Math} & Numerical premise & Recompute with the revised value \\
\bottomrule
\end{tabularx}
\end{table}

\vspace{2mm}
\paragraph{\domainlabel{Actions}.}
\label{app:construction_actions}
The 104 Actions revisions change one tool argument in a call or a specified group of calls. Each proposal identifies the affected calls.
We check the edited argument's type and allowed values against the tool
schema. The updated calls must also satisfy any total specified in the task.
We exclude changes to shared settings, such as units, that would alter the
meaning of other argument values.
When several targets are valid, we prefer one introduced before the final
source passage so that task information remains to be revealed after the event.

After updating the affected reference arguments, we check both task states.
Each reference must pass its own verifier and fail the verifier for the other
state. To score a response, we parse it into function calls and compare their
structure and argument values with those allowed by the active task.

\paragraph{Example.}
In \texttt{sharded-BFCL/parallel\_143}, the user asks for the total area of
three triangular gardens. The base and height pairs are $(10,5)$, $(15,7)$,
and $(20,10)$ meters. The proposal and the affected reference call are shown
below:

\begingroup
\tcbset{promptstyle/.append style={colback=domainActions!5!white,
  colbacktitle=domainActions,coltitle=white,colframe=domainActions}}
\begin{promptbox}{Actions: proposal and affected reference call}
\textbf{Proposal:} For the \texttt{calc\_area\_triangle} call with base 10,
use 6 for the \texttt{height} argument. Would this be suitable?\par\smallskip
\textbf{Original:} \texttt{calc\_area\_triangle(base=10, height=5)}\par
\textbf{Revised:} \texttt{calc\_area\_triangle(base=10, height=6)}
\end{promptbox}
\endgroup

\textbf{The selected reference call uses \texttt{height=6} only if the
proposal is accepted. Otherwise, it keeps \texttt{height=5}.}
The other two calls, for pairs $(15,7)$ and $(20,10)$, stay unchanged.

\vspace{2mm}

\paragraph{\domainlabel{Code}.}
\label{app:construction_code}
The 100 Code revisions change the required entry-point function or method name recorded in the task metadata. The original name appears in the first source message, and a later proposal specifies the new name.

We update the name in the task contract and any available reference code, leaving the computation unchanged. Both task states use the same verifier and execution tests, with different naming requirements. When both reference programs are available, each must pass under its corresponding task state and fail under the other. Otherwise, we check that the specified name change has been applied.

To evaluate generated programs, we check their entry-point names and run the execution tests. In both single- and multi-turn conditions, a program must use the new name if the proposal is accepted and the original name otherwise. Defining both names is not allowed.

\paragraph{Example.}
In \texttt{sharded-HumanEval/5}, the task inserts a delimiter between adjacent
numbers in a list. The box shows the proposal to rename the function,
followed by a test call to each reference program:

\begingroup
\tcbset{promptstyle/.append style={colback=domainCode!5!white,
  colbacktitle=domainCode,coltitle=white,colframe=domainCode}}
\begin{promptbox}{Code: proposal, reference calls, and shared output}
\textbf{Proposal:} Use \texttt{intersperse\_revised} as the required function
name. Would you recommend this option?\par\smallskip
\textbf{Original:} \texttt{intersperse([5, 6, 3, 2], 8)}\par
\textbf{Revised:} \texttt{intersperse\_revised([5, 6, 3, 2], 8)}\par
\textbf{Result in both cases:} \texttt{[5, 8, 6, 8, 3, 8, 2]}
\end{promptbox}
\endgroup

\textbf{The required function name changes only if the proposal is accepted.
Otherwise, it remains \texttt{intersperse}.}
The function body and expected outputs remain unchanged.

\vspace{2mm}

\paragraph{\domainlabel{Database}.}
\label{app:construction_database}
We select a query requirement in the original single-turn task. A candidate
proposes a new value for that requirement and updates the corresponding SQL
condition. We also consider changes to ordering, set
membership, comparisons, aggregation, and output fields, in that order.
If the initial rules yield no valid candidate, we keep the first valid
candidate from these additional edits.
We run the original and revised reference queries on the task's database.
If they return the same result, we discard the revision.
The 107 events comprise 88 SQL-condition changes and 19 output-field changes.

Generated queries are scored by comparing their execution results with the
active reference on that database instance.

\paragraph{Example.}
In \texttt{sharded-spider-val-265-medium}, the user asks for the cities of
origin of employees under age 30, keeping only cities represented by more
than one qualifying employee. The proposal and the two reference queries
are shown below:

\begingroup
\tcbset{promptstyle/.append style={colback=domainDatabase!5!white,
  colbacktitle=domainDatabase,coltitle=white,colframe=domainDatabase}}
\begin{prompttemplate}{Database: proposal and reference queries}
Proposal:
the employees in question should be under the age of 32. How would this revised requirement work?

Original:
SELECT city FROM employee WHERE age < 30
GROUP BY city HAVING count(*) > 1

Revised:
SELECT city FROM employee WHERE age < 32
GROUP BY city HAVING count(*) > 1
\end{prompttemplate}
\endgroup

Only the age cutoff changes, from 30 to 32.
\textbf{On the task's database, the reference result changes to Bath and
Bristol only if the proposal is accepted. Otherwise, it remains Bath.}

\vspace{2mm}

\paragraph{\domainlabel{Math}.}
\label{app:construction_math}
For each of the 103 Math tasks, we replace one numerical premise in the
original single-turn task. For rule-based edits, the selected passage must
contain a single numerical target whose value does not appear in other
passages. We exclude numbers used as labels and replacements that conflict
with other quantities.

We recompute the revised answer and check that both reference answers are
valid and different. To score a response, we extract its final number,
normalize its format, and compare it with the reference for the active task.

\paragraph{Example.}
In \texttt{sharded-GSM8K/799}, bread costs \$2 per loaf and bagels cost \$1
each. The task asks how much more three loaves cost than two bagels.

\begingroup
\tcbset{promptstyle/.append style={colback=domainMath!5!white,
  colbacktitle=domainMath,coltitle=white,colframe=domainMath}}
\begin{tcolorbox}[promptstyle,unbreakable,title={Math: proposal and reference answers}]
\textbf{Proposal:} each bagel costs \$2. Would this be a suitable update?\par\smallskip
\textbf{Original:} $3\times\$2-2\times\$1=\$4$\par
\textbf{Revised:} $3\times\$2-2\times\$2=\$2$
\end{tcolorbox}
\endgroup

\textbf{The reference answer changes to \$2 only if the proposal is accepted.
Otherwise, it remains \$4.}

\paragraph{Selecting a revision.}
When several rule-based candidates remain for a task, we first choose the
requirement to change. We apply the domain-specific priorities above, then
favor longer source passages and use earlier positions to break ties.
For that requirement, a fixed rule based on the source task ID selects one
valid replacement. Selection is independent of model outputs.

\paragraph{Human review and condition rendering.}
We review each selected revision alongside its source task to check that the
change is clear and consistent with the other requirements. We use LLMs to
draft templates with varied wording for proposals, decisions, and clarification
questions. After human review, we use these templates to render approved
revisions into dialogue messages. We save the target passage, resulting
messages, and revised reference in one record.

Retained and Revised share the proposal, paired with rejection or acceptance,
respectively. Neutral uses two template-based clarification questions about
the same requirement, and Original has no inserted event. Where multiple
template variants are available, a stable rule based on the record identifier
selects the wording.

We insert the event after the message that introduces the target requirement, then continue
with any remaining task information. Each condition uses the reference
specified in Section~\ref{sec:conditions}.

Neutral questions refer to the requirement being revised. Automated checks
confirm that they reveal neither the original nor the replacement value,
that Retained and Revised share the same proposal, and that decision messages
do not repeat the proposed value. All 414 selected revisions pass these checks.

\subsection{A complete worked example}
\label{app:worked_example}
We now walk through one evaluated task. The onion-purchase task in
\texttt{sharded-GSM8K/248}, shown in Figure~\ref{fig:construction}, provides
the source prompt and four shards below. All conditions share the same

\vspace{2mm}
\domainname{Math} system message.

\begingroup
\tcbset{promptstyle/.append style={colback=domainMath!5!white,
  colbacktitle=domainMath,coltitle=white,colframe=domainMath}}
\begin{prompttemplate}{Math System Message}
As an expert problem solver solve step by step the following mathematical questions.
\end{prompttemplate}

\begin{prompttemplate}{Original Single-Turn Prompt ($S_{\mathrm{Orig}}$)}
Q: A chef bought (*@\intentblue{\textbf{4 bags}}@*) of onions. Each bag weighs 50 pounds. A pound of onions cost (*@\textdollar@*)1.50. How much did the chef spend?
A:
\end{prompttemplate}

\begin{promptbox}{Source Shards}
\textbf{S1:} calculate how much the chef spent on onions\par
\textbf{S2:} the chef bought \intentblue{\textbf{4 bags}} of onions\par
\textbf{S3:} each of those bags weighs 50 pounds\par
\textbf{S4:} and each pound of onions costs \$1.50
\end{promptbox}
\vspace{2mm}

\paragraph{1. Selecting the target and updating the reference.}
The selected target is the \textbf{bag count} in S2, which changes from \intentblue{\textbf{four}} to \intentorange{\textbf{two}}.
The weight per bag, price per pound, and spending question remain unchanged.
Recomputing the cost gives
\[
y^{(0)}=\intentblue{\mathbf{4}}\times50\times\$1.50=\intentblue{\$\mathbf{300}},
\qquad
y^{(1)}=\intentorange{\mathbf{2}}\times50\times\$1.50=\intentorange{\$\mathbf{150}}.
\]
\paragraph{2. Placing the event in the dialogue.}
Each labeled item is sent as a separate user message. The labels are
included here only to guide the reader.

\vspace{2mm}
\begin{promptbox}{Proposal and Decisions for This Example}
\textbf{Propose (shared):} the chef bought \intentorange{\textbf{2 bags}} of onions. Does this alternative
make sense?\par
\textbf{Reject (Retained):} Ignore that revision.\par
\textbf{Accept (Revised):} Let it change as proposed.
\end{promptbox}

\textbf{The reference answer changes to \intentorange{\$150} only if the proposal is
accepted. Otherwise, it remains \intentblue{\$300}.}

\begin{promptbox}{Canonical Neutral Messages}
\textbf{Neutral 1:} How does the \textbf{number of onion bags} the chef bought enter the calculation?

\textbf{Neutral 2:} Why is the \textbf{number of onion bags} the chef bought important for determining
the requested result?
\end{promptbox}
\vspace{2mm}

The event follows S2, once the bag count has been introduced. S3 and S4 then
supply the weight and price needed for the calculation, as shown in
Figure~\ref{fig:construction}. The user's decision determines which bag count
is active (Table~\ref{tab:construction_example_targets}). The post-decision
Neutral suffix reuses the two clarification questions above.

\begin{table}[htbp]
\centering
\small
\vspace{2mm}
\caption{Condition-specific targets for the worked Math example.}
\vspace{2mm}
\label{tab:construction_example_targets}
\begin{tabular}{@{}llll@{}}
\toprule
Condition & Inserted event & Active bag count & Reference answer \\
\midrule
Original & None & \intentblue{\textbf{4}} & \intentblue{\textbf{\$300}} \\
Neutral & Clarification questions & \intentblue{\textbf{4}} & \intentblue{\textbf{\$300}} \\
Retained & Propose \intentorange{\textbf{2 bags}}, then reject & \intentblue{\textbf{4}} & \intentblue{\textbf{\$300}} \\
Revised & Propose \intentorange{\textbf{2 bags}}, then accept & \intentorange{\textbf{2}} & \intentorange{\textbf{\$150}} \\
\bottomrule
\end{tabular}
\vspace{2mm}
\end{table}

In evaluation, each conversation is generated independently using the
event messages reproduced above.

\paragraph{3. Single-turn controls.}
Each box contains one complete user message, including the
connectors that introduce the proposal and decision.

\begin{prompttemplate}{Direct Revised Single-Turn Prompt ($S_{\mathrm{Dir}}$)}
Q: A chef bought (*@\intentorange{\textbf{2 bags}}@*) of onions. Each bag weighs 50 pounds. A pound of onions cost (*@\textdollar@*)1.50. How much did the chef spend?
A:
\end{prompttemplate}

\begin{prompttemplate}{Revised Single-Turn Prompt ($S_{\mathrm{Rev}}$)}
Q: A chef bought (*@\intentblue{\textbf{4 bags}}@*) of onions. Each bag weighs 50 pounds. A pound of onions cost (*@\textdollar@*)1.50. How much did the chef spend? Before you go, I have one quick last minute thought: The chef bought (*@\intentorange{\textbf{2 bags}}@*) of onions. (*@\textbf{Okay, let's do that.}@*)
A:
\end{prompttemplate}

\begin{prompttemplate}{Retained Single-Turn Prompt ($S_{\mathrm{Ret}}$)}
Q: A chef bought (*@\intentblue{\textbf{4 bags}}@*) of onions. Each bag weighs 50 pounds. A pound of onions cost (*@\textdollar@*)1.50. How much did the chef spend? Before you go, I have one quick last minute thought: The chef bought (*@\intentorange{\textbf{2 bags}}@*) of onions. (*@\textbf{Wait, that does not seem quite right. Ignore that revision.}@*)
A:
\end{prompttemplate}
\endgroup
\vspace{2mm}

%% file: appendix/c_evaluation.tex
\section{Evaluation Settings and User Simulation}
\label{app:evaluation_settings}

\subsection{Models and generation settings}
\paragraph{Model versions.}
For Llama-3.1-8B, we use the Unsloth distribution of
Meta-Llama-3.1-8B-Instruct. The Gemma model is Gemma-4-26B-A4B-it.

\begin{table}[htbp]
\centering\small
\vspace{2mm}
\caption{Generation settings for local models and the user simulator.
Output limits are in tokens per response.}
\vspace{2mm}
\label{tab:evaluation_settings}
\begin{tabular*}{\linewidth}{@{\extracolsep{\fill}}lcc@{}}
\toprule
Setting & Local single-turn & Local multi-turn \\
\midrule
Output limit & 1,000 & 4,096 \\
Temperature & 0 & 0 \\
Top-$p$ & 1 & 1 \\
Seed & 0 & Derived from seed 0 \\
\midrule
\multicolumn{3}{@{}l}{\textbf{User simulator (all multi-turn runs)}} \\
Model & \multicolumn{2}{l@{}}{Qwen3.6-27B} \\
Temperature & \multicolumn{2}{l@{}}{0.3} \\
Output limit & \multicolumn{2}{l@{}}{192} \\
\bottomrule
\end{tabular*}
\vspace{2mm}
\end{table}

API models use default sampling settings. Their single-turn and multi-turn output limits
match those in Table~\ref{tab:evaluation_settings}, including reasoning tokens.

\paragraph{Scoring and aggregation.}
The domain verifier in Appendix~\ref{app:construction_scoring} scores the final eligible
response after source disclosure and the event path are complete. Benchmark
accuracy pools sources within each model and then averages models with equal
weight.

\subsection{User simulation}
\label{app:user_simulation}
Following LiC~\citep{laban2026lost}, a \textit{shard} is a short text
fragment containing part of a task, such as its goal or one requirement.
To \textit{reveal} a shard is to convey its information to the assistant
in a user message. The \textit{user simulator} is the LLM used to generate
user messages, rephrasing shards to fit the ongoing dialogue.
The four multi-turn conditions are matched by source task.
The evaluation protocol directly inserts the first shard.
The user simulator then generates messages based on the evolving
dialogue history. After the relevant requirement has been introduced,
the protocol inserts the fixed \textit{event messages} specified for
that condition, including clarifications, proposed changes, and
acceptance or rejection messages.

An \textit{eligible shard} has not yet been revealed and is allowed by the
protocol on the current turn. After the inserted event in
\textit{Neutral}, \textit{Retained}, or \textit{Revised}, the simulator must
resume task disclosure by revealing an eligible shard. We call this a
\textit{mandatory-disclosure turn}. The script checks the JSON reply's
format, selected shard ID, and text, rejecting empty replies and bare
acknowledgments. If a check fails, it uses a \textit{fallback}: the stored
text of the first eligible shard is sent unchanged. When consecutive
replies reveal no new shard, a recovery step likewise requires the
simulator to disclose an eligible shard.

\begingroup
\tcbset{breakable}
\input{appendix/c_prompts}
\endgroup

\subsection{User simulator reliability}
\label{app:simulator_reliability}
We evaluated the LLM-based user simulator, including its fallback and
recovery mechanisms. A manual inspection covered
\textbf{2,345 simulated user turns} across 500 recorded conversations,
with 125 conversations from each of \domainname{Actions},
\domainname{Code}, \domainname{Math}, and \domainname{Database}.
We adapted the user-turn inspection dimensions of \citet{laban2026lost}
to assess messages actually delivered to the assistant.

\textbf{Task-relevant information delivery} checks whether a shard-bearing
message conveys its task-relevant content in the context available at
that turn. Paraphrasing, omission of task-irrelevant background, and faithful
combinations of shards are accepted. This criterion was met in
\textbf{2,284 of 2,317 shard-bearing messages (98.6\%)}.
\textbf{Contextualization} checks whether a message appropriately addresses
the preceding assistant response. All \textbf{2,345 inspected messages (100\%)}
were judged appropriately contextualized.

%% file: appendix/c_prompts.tex
For each simulator call, the evaluation script fills the following inputs:
\begin{itemize}[leftmargin=*,itemsep=2pt,parsep=0pt]
\item \texttt{[[CONVERSATION\_SO\_FAR]]}: the user and assistant messages so
far, excluding the assistant's system instructions.
\item \texttt{[[SHARDS\_REVEALED]]}: shards already supplied in user messages,
listed with their IDs and text.
\item \texttt{[[SHARDS\_NOT\_REVEALED]]}: shards not yet supplied and available
on this turn. Shards held back for later turns are excluded.
\end{itemize}
The JSON reply has two fields: \texttt{response} contains the next user
message, and \texttt{shard\_id} identifies the shard it reveals ($-1$ when
none is revealed). Only \texttt{response} is sent to the evaluated model.

\vspace{2mm}
\begin{prompttemplate}{User Simulator Prompt (LiC-Derived)}
You are simulating a user of an interactive LLM system (like ChatGPT).
The user is inherently lazy, and answers in short form, providing only minimal information to the system. You should not be proactive.

Here's the conversation so far:
[[CONVERSATION_SO_FAR]]

Here are the shards that have already been revealed:
[[SHARDS_REVEALED]]

Here are all the shards that have not been revealed yet:
[[SHARDS_NOT_REVEALED]]

You must generate a response to the conversation so far. Here are the rules:
- [Providing a Shard] You can reveal the content of a shard to the system in your response if it will help the system move closer to answering the problem. You should select the shard to reveal that is most "basic" and is the current most relevant shard.
- [One Shard at a Time] You should only reveal at most one shard at a time.
- [Reveal Entire Shard] If you reveal a shard, you must make sure to include *all the information in the shard*. For example, if the shard is "your symptoms are that you have a headache in the mornings", your response can't just be ``yeah I have headaches'', you must say ``yup mostly headaches in the mornings``.
- [Irrelevant Clarifications] If the system asks you a question irrelevant to the shards, asks you a generic question (``Can you give me a hint?``), you should respond with an answer that does not provide a shard. (``I don't know``, ``Is that really important?``, etc.) You should not reveal any information beyond what is available in the shards.
- [No Repeated Shards] You should not reveal the same shard more than once. Carefully review the shards revealed already, and only reveal a shard if its `shard_id` is not on the list.
- [Rephrase Shards] If you reveal a shard, you should rephrase it in a conversational way. Do not copy the shard verbatim.
- [Do Not Ask Questions] Your response should always be declarative sentences, and not questions.
- [Brevity of Response] You should favor being succint. Your answer can also have typos, improper grammar, capitalization, etc. You are simulating a real person talking to an AI, who is in a hurry.
- [Task Changes] Task changes and accept/reject decisions are supplied separately. Do not introduce or reverse them, or restore source values superseded by an accepted change.
- Use only facts from the source shards. Ignore anything the Assistant calculated, guessed, or answered; do not repeat or confirm it as a user fact.
- [Format] Your response should be formatted as a JSON object with the following keys:
    - `response`: The response to the conversation so far.
    - `shard_id`: The shard you are revealing to the system. The shard_id can be an integer, or -1 if you did not reveal any shards.
For example:
{"response": "I don't know", "shard_id": -1}
or:
{"response": "yeah I want it to [...]", "shard_id": 1}
\end{prompttemplate}
\vspace{2mm}

On turns that must reveal new task information, the evaluation script adds
the following instruction:

\vspace{2mm}
\begin{prompttemplate}{Mandatory-Disclosure Instruction}
Current turn requirement: reveal exactly one unrevealed shard. Return its shard_id, not -1; do not give only an acknowledgement.
\end{prompttemplate}
\vspace{2mm}

%% file: appendix/b_conditions.tex
\section{Interaction Extension Protocols}
\label{app:interaction_protocols}

Section~\ref{sec:single_controls} defines the single-turn controls;
Appendices~\ref{app:worked_example} and~\ref{app:additional_model_results}
provide a complete prompt example and numerical comparisons, respectively.

\subsection{Accumulation: repeated clarification or rejection}
We write the $\ell$th clarification block as
$N^{(\ell)}=[n^{(\ell)}_1,n^{(\ell)}_2]$, which contains two Neutral messages
about the same selected requirement, without introducing a replacement value.
The corresponding Retained block is
$P^{(\ell)}=[\textsc{Propose}(r'_\ell),\textsc{Reject}]$.
Each block proposes a different replacement for that requirement and then
rejects it. Thus, the requirement being discussed stays fixed across blocks,
and the original requirement remains active after every rejection.

The operator $\diamond$ joins the blocks in order. Each event message is sent
as a separate user turn, followed by an assistant reply. The operator
$\oplus_j$ inserts the sequence after the assistant's reply to $u_j$, the
user message that first introduces the selected requirement, and before
the remaining task information is disclosed. With $k$ blocks, the two paths are
\begin{equation}
\begin{aligned}
M_{\mathrm{Neu}}^{(k)}&=M_{\mathrm{Orig}}\oplus_j
 (N^{(1)}\diamond\cdots\diamond N^{(k)}),\\
M_{\mathrm{Ret}}^{(k)}&=M_{\mathrm{Orig}}\oplus_j
 (P^{(1)}\diamond\cdots\diamond P^{(k)}),\\
M_{\mathrm{Neu}}^{(0)}&=M_{\mathrm{Ret}}^{(0)}=M_{\mathrm{Orig}}.
\end{aligned}
\label{eq:accumulation_protocol}
\end{equation}
Both paths add $2k$ event user messages.

\subsection{Persistence: clarification after a decision}
For $v\in\{\mathrm{Ret},\mathrm{Rev}\}$, let
$D_v=[\textsc{Propose}(r'),\textsc{Decision}_v]$ denote the canonical
proposal followed by rejection or acceptance, sent as two separate user
messages and held fixed across depths. The $k$ Neutral blocks begin after the assistant's reply to the
decision message, before any remaining task information is disclosed:
\begin{equation}
\begin{aligned}
M_{v\to\mathrm{Neu}}^{(k)}
  &=M_{\mathrm{Orig}}\oplus_j
    (D_v\diamond N^{(1)}\diamond\cdots\diamond N^{(k)}),\\
M_{v\to\mathrm{Neu}}^{(0)}&=M_v.
\end{aligned}
\label{eq:persistence_protocol}
\end{equation}
The same clarification questions are used after either decision. They usually
concern the selected requirement without proposing new values.
These questions do not change the active task: Retained keeps the original
requirement active, while Revised keeps the accepted revision active at every depth.

%% file: appendix/d_training.tex
\section{\method{} Training and Evaluation}
\label{app:opsd_training}

\paragraph{Active-task reference construction.}
FiC~\citep{chen2026found} constructs its Teacher input from concatenated
shards, while MAIGO~\citep{zheng2026maigo} removes assistant replies and adds
a complete-task reference at the answer turn. Applying these constructions
here requires specifying which requirements remain active, since user
messages can also contain rejected or superseded content. \method{}
explicitly aligns supervision with the user's decision: rejection selects
the original Teacher task, while acceptance selects the revised task.
The Student receives the full dialogue and is trained to follow the
resulting active requirements.

\subsection{Training data and conversation construction}

Training, validation, and evaluation use disjoint source IDs.
We provide 50 validation sources per domain.
The \domainname{Actions}, \domainname{Database}, and \domainname{Math}
training pools derive from BFCL~\citep{patil2025bfcl},
Spider~\citep{yu2018spider}, and GSM8K~\citep{cobbe2021training}, respectively.
The \domainname{Code} pool contains 193 MBPP~\citep{austin2021program}
tasks, 56 HumanEval~\citep{chen2021evaluating} tasks, and 96 easy and 155
medium LiveCodeBench~\citep{jain2025livecodebench} tasks.

We initially assemble 500 training candidates per domain and apply
domain-specific data checks. For \domainname{Database}, we check that the original and revised
queries match their respective task requirements and produce different
results on the task database. These checks leave 381
\domainname{Database} candidates.
The resulting pools then undergo the Teacher admission check in
Section~\ref{sec:method_insight}, using a
frozen copy of the Student's base model as the Teacher.
Table~\ref{tab:training_sources} reports the resulting training counts
for each model and domain.

\begin{table}[htbp]
\centering\small
\vspace{2mm}
\caption{Teacher-admitted training sources by model and domain. Each source
contributes all four conditions.}
\vspace{2mm}
\label{tab:training_sources}
\begin{tabular}{@{}lcccc@{}}
\toprule
Model & \domainlabel{Actions} & \domainlabel{Code} & \domainlabel{Database} & \domainlabel{Math} \\
\midrule
Qwen3-8B & 435 & 316 & 201 & 451 \\
Gemma-4-26B-A4B & 447 & 367 & 307 & 428 \\
Qwen2.5-7B-Instruct & 323 & 322 & 214 & 454 \\
Llama-3.1-8B & 407 & 266 & 212 & 455 \\
\bottomrule
\end{tabular}
\vspace{2mm}
\end{table}

Training conversations follow the sharded conversation setup of
LiC~\citep{laban2026lost}. Each complete task is divided into information
shards, and a user simulator reveals at most one shard per turn while
interacting with the current Student. We add the benchmark's
condition-specific clarification, proposal, and decision messages to these
conversations. The resulting histories are generated online as the Student
changes during training.

\subsection{Training objectives and generation budgets}
Both \method{} and matched SFT use LoRA~\citep{hu2021lora}, with shared
training hyperparameters reported in Table~\ref{tab:training_hyperparameters}.

\paragraph{\method{}.}
Training uses the final-answer distillation objective in
Equation~\ref{eq:opsd_loss}. Correct and incorrect Student answers both
contribute to the loss, which is averaged over conditions and sources.

\paragraph{Matched SFT.}
SFT reuses the selected sources and dialogue histories $H_v$ recorded during
\method{} training.
For each condition $v$, its target $y_v^{\mathrm{T}}$ is the verified Teacher answer
recorded during admission for the corresponding complete task $c_v$
(Equation~\ref{eq:teacher_view}). For each source, SFT minimizes
\begin{equation}
\mathcal{L}_{\mathrm{SFT}}(\theta)
= -\frac{1}{4}\sum_v
\frac{1}{|y_v^{\mathrm{T}}|}
\sum_{t=1}^{|y_v^{\mathrm{T}}|}
\log \pi_\theta\!\left(
 y_{v,t}^{\mathrm{T}} \mid H_v,y_{v,<t}^{\mathrm{T}}
\right).
\label{eq:matched_sft}
\end{equation}
Here, $v$ ranges over the four conditions. The loss is averaged over sources,
and history tokens provide context without receiving supervision.
Both methods start from the same base model and use the same training
sources, condition weights, and optimization schedule.
Both methods receive supervision aligned with the user's final decision.
We compare on-policy distribution matching with hard-target learning
under shared active-task supervision.

\begin{table}[htbp]
\centering\small
\caption{Training hyperparameters shared by \method{} and matched SFT
within each model--domain pair. Each source contributes all four conditions.}
\label{tab:training_hyperparameters}
\begin{tabular*}{\linewidth}{@{\extracolsep{\fill}}ll@{}}
\toprule
Hyperparameter & Setting \\
\midrule
Optimizer & AdamW \\
Learning rate & $1\times10^{-5}$ \\
Weight decay & 0 \\
Learning-rate schedule & Constant, without warmup \\
Training epochs & 1 \\
Sources per optimizer update & 4 (Actions, Database); 8 (Code, Math) \\
LoRA rank / alpha / dropout & 16 / 32 / 0 \\
Training seed & 42 \\
\bottomrule
\end{tabular*}
\end{table}

\paragraph{Training budgets.}
For \domainname{Code}, \domainname{Database}, and \domainname{Math}, the
maximum training sequence length is 32,768 tokens, with up to 4,096 tokens
for both intermediate replies and final answers. \domainname{Actions} uses
8,192 tokens per training sequence, 128 per intermediate reply, and 512 per
final answer. Its output format requires function calls without explanatory
text, motivating a more compact training budget. Evaluation uses the common
multi-turn generation settings in Appendix~\ref{app:evaluation_settings}.

%% file: appendix/e_supplementary_results.tex
\section{Supplementary Results and Error Analysis}
\label{app:supplementary_results}

We follow the research questions in order: additional models and single-turn
controls for RQ1, error patterns for RQ2, interaction-depth analysis for
RQ3, and distillation results across depths for RQ4.

\subsection{Overall performance (RQ1)}
\label{app:additional_model_results}
\paragraph{Additional model results.}
Table~\ref{tab:supplementary_models} provides supplementary results for
Qwen3-4B-Instruct-2507 and Qwen2.5-7B-Instruct under the same conditions as
Table~\ref{tab:main_results}. These two models are not included in the
eight-model Overall aggregate in Table~\ref{tab:main_results}.
In every domain, both models score lower on Multi Original than on Single
Original, and lower again on Multi Retained.

\input{tables/supplementary_models}
\FloatBarrier

\subsection{Error diagnostics and examples (RQ2)}
\label{app:error_analysis}
We analyze the multi-turn outputs in Tables~\ref{tab:main_results} and
\ref{tab:supplementary_models}, selecting cases scored correctly in Original but
incorrectly in Retained or Revised. \textit{Inactive content} refers to the rejected
proposal in Retained or the original requirement replaced in Revised.
We apply domain-specific automatic checks to the complete final response:
argument values in \domainname{Actions}, entry-point identifiers in
\domainname{Code}, SQL literals, operators, or predicates in
\domainname{Database}, and premise or reference-answer values in
\domainname{Math}. Explanations and code comments are included.
We report the fraction of selected error responses that contain
inactive content.

\begin{figure}[htbp!]
  \centering
  \vspace{-2mm}
  \includegraphics[width=0.8\textwidth]{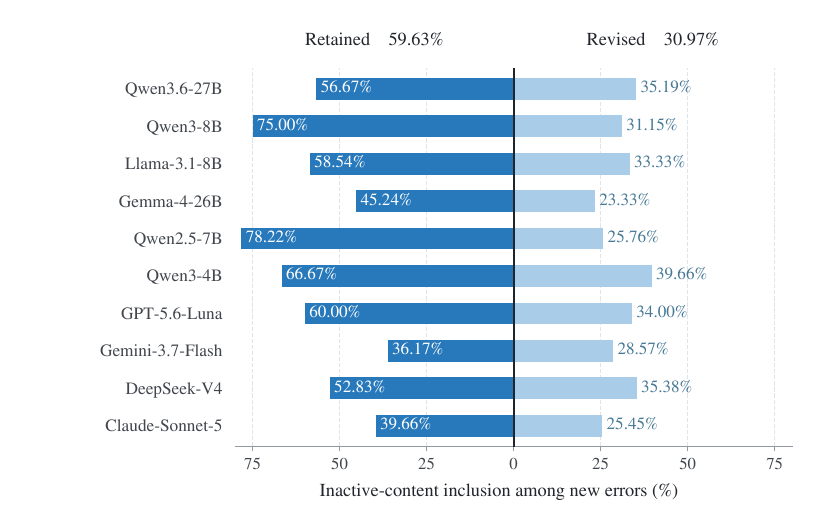}
  \vspace{-4mm}
  \caption{Inactive content appears more often in Retained errors for all 10
  models. Bars show multi-turn content-inclusion rates among Original-correct
  outputs that become incorrect in the named condition. Headers pool all 10
  models.}
  \label{fig:inactive_content_by_model}
\end{figure}

\paragraph{Multi-turn results.}
Across all ten models, inactive content appears in 446 of 748 new Retained
errors (59.63\%) and 201 of 649 new Revised errors (30.97\%).
Figure~\ref{fig:inactive_content_by_model} provides the model-level breakdown.

\paragraph{Examples.}
\label{app:multi_turn_error_examples}
The examples below show rejected or superseded values used in calculations
that produce incorrect final answers.

\vspace{2mm}
\begingroup
\tcbset{promptstyle/.append style={colback=domainMath!5!white,
  colbacktitle=domainMath,coltitle=white,colframe=domainMath}}

\begin{tcolorbox}[promptstyle,unbreakable,title={Example 1: using a rejected value}]
\textbf{Model:} GPT-5.6-Luna, multi-turn Retained.\par
\textbf{Source:} \texttt{sharded-GSM8K/543}.\par
\textbf{Task (summary):} Jairus earns \intentblue{\textbf{\$0.80 per task}}
and Jenny earns \$0.50. Each completes 20 tasks. How much more does Jairus
earn?\par
\textbf{Proposed change (summary):} Jairus earns
\intentorange{\textbf{\$0.90 per task}}.\par
\textbf{User decision:} ``That revision does not work for me.''\par
\textbf{Active answer:} \intentblue{\textbf{\$6}}.\par
\textbf{Response (summary):} Uses \intentorange{\textbf{\$0.90 per task}} and returns
\intentorange{\textbf{\$8}}.
\end{tcolorbox}
\vspace{2mm}

\phantomsection
\label{app:revised_error_example}
\begin{tcolorbox}[promptstyle,unbreakable,title={Example 2: keeping the original value after acceptance}]
\textbf{Model:} Claude-Sonnet-5, multi-turn Revised.\par
\textbf{Source:} \texttt{sharded-GSM8K/698}.\par
\textbf{Task (summary):} Greta receives a \$2,400 paycheck, puts 50\% into
retirement, and spends \intentblue{\textbf{20\% on her car payment}}.
How much remains?\par
\textbf{Proposed change (summary):} Spend
\intentorange{\textbf{21\% on the car payment}}.\par
\textbf{User decision:} ``Authorize that adjustment.''\par
\textbf{Active answer:} \intentorange{\textbf{\$696}}.\par
\textbf{Response (summary):} Lists the car payment as
\intentblue{\textbf{20\% (\$480)}} and reports
\intentblue{\textbf{\$720}} remaining.
\end{tcolorbox}

\endgroup

\subsection{Interaction depth in six additional models (RQ3)}
\label{app:interaction_depth_results}

\begin{figure}[H]
  \centering
  \includegraphics[width=1.0\textwidth,trim=2bp 5bp 2bp 5bp,clip]{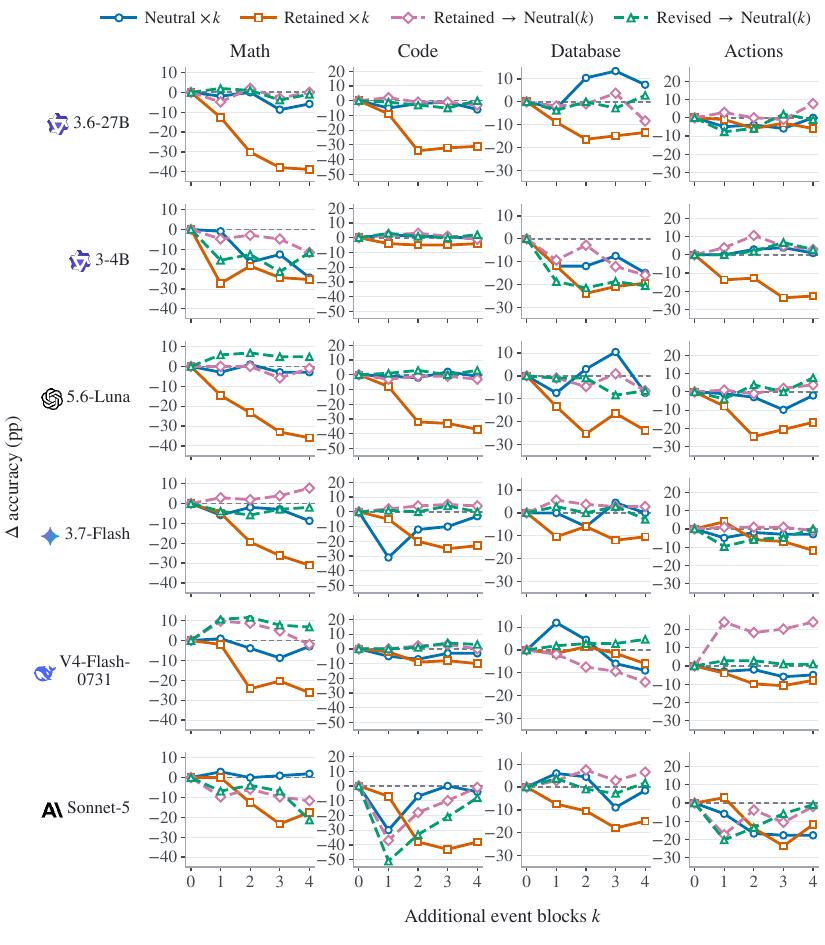}
  \captionsetup{skip=4pt}
  \vspace{-5mm}
  \caption{Interaction depth in six additional models. Accuracy changes are
  relative to each curve's baseline at $k=0$.}
  \label{fig:supplementary_interaction_depth}
  \vspace{-2mm}
\end{figure}

Figure~\ref{fig:supplementary_interaction_depth} complements
Section~\ref{sec:suffix_results} with six additional models evaluated on the
same 372-source accumulation and 414-source persistence cohorts.

These results support the main-text trends. Averaged across the six models
and four interaction paths, the first added block produces the largest
single-step accuracy drop. Repeated Retained produces larger average losses
than Neutral, with pronounced declines in Math. After either decision,
additional clarification also leaves mean accuracy below the corresponding
baseline at the final depth, with effects varying across models and domains.

\FloatBarrier
\subsection{Intent-OPSD across interaction depths (RQ4)}
\label{app:opsd_depth_results}
We evaluate the four RQ4 models on 372 accumulation and 414 persistence
sources per model, pooling counts within each path and depth
(Table~\ref{tab:opsd_depth}).
Accumulation starts from Original; persistence starts from the respective
decision baseline. Code inputs specify the original entry-point name
before revision proposals.

\begin{table}[htbp]
\centering
\vspace{2mm}
\caption{Base, matched SFT, and \method{} accuracy (\%) across interaction depths.
Each path pools 1,488 evaluations for accumulation or 1,656 for persistence;
Mean pools all four paths at each depth.
Colors follow Table~\ref{tab:method_results}, comparing each method with Base at the same depth.}
\vspace{2mm}
\label{tab:opsd_depth}
\definecolor{depthGain}{HTML}{90EE90}
\definecolor{depthTie}{HTML}{FFFFE0}
\definecolor{depthLoss}{HTML}{e66565}
\small
\begin{tabular}{llrrrrr}
\toprule
Interaction path & Method & $k=0$ & $k=1$ & $k=2$ & $k=3$ & $k=4$ \\
\midrule
Accumulation: Neutral & Base & 47.58 & 43.95 & 41.53 & 40.12 & 39.45 \\
 & SFT & \cellcolor{depthGain!25.01}55.98 & \cellcolor{depthGain!26.90}52.69 & \cellcolor{depthGain!32.42}52.49 & \cellcolor{depthGain!34.92}51.88 & \cellcolor{depthGain!37.66}52.28 \\
 & \method{} & \cellcolor{depthGain!33.29}60.62 & \cellcolor{depthGain!34.82}56.79 & \cellcolor{depthGain!43.42}57.86 & \cellcolor{depthGain!45.03}56.65 & \cellcolor{depthGain!47.79}56.99 \\
\midrule
Accumulation: Retained & Base & 47.58 & 32.19 & 27.08 & 25.87 & 26.81 \\
 & SFT & \cellcolor{depthGain!25.01}55.98 & \cellcolor{depthGain!31.47}40.32 & \cellcolor{depthGain!44.80}38.17 & \cellcolor{depthGain!47.97}37.43 & \cellcolor{depthGain!47.28}38.58 \\
 & \method{} & \cellcolor{depthGain!33.29}60.62 & \cellcolor{depthGain!27.57}38.84 & \cellcolor{depthGain!42.48}37.43 & \cellcolor{depthGain!42.90}35.89 & \cellcolor{depthGain!36.63}35.22 \\
\midrule
Persistence: Retained & Base & 31.40 & 30.07 & 28.56 & 27.78 & 28.68 \\
 & SFT & \cellcolor{depthGain!30.76}39.07 & \cellcolor{depthGain!30.31}37.26 & \cellcolor{depthGain!29.77}35.21 & \cellcolor{depthGain!33.65}35.51 & \cellcolor{depthGain!27.36}34.54 \\
 & \method{} & \cellcolor{depthGain!29.45}38.59 & \cellcolor{depthGain!32.02}37.86 & \cellcolor{depthGain!32.64}36.17 & \cellcolor{depthGain!32.36}35.08 & \cellcolor{depthGain!28.07}34.78 \\
\midrule
Persistence: Revised & Base & 40.88 & 38.04 & 37.26 & 35.27 & 37.38 \\
 & SFT & \cellcolor{depthGain!14.52}43.06 & \cellcolor{depthGain!16.88}41.12 & \cellcolor{depthGain!17.85}40.70 & \cellcolor{depthGain!24.85}41.43 & \cellcolor{depthGain!19.61}41.61 \\
 & \method{} & \cellcolor{depthGain!30.97}50.97 & \cellcolor{depthGain!34.96}49.21 & \cellcolor{depthGain!39.62}50.24 & \cellcolor{depthGain!41.73}48.43 & \cellcolor{depthGain!34.99}48.37 \\
\midrule
Mean & Base & 41.56 & 35.96 & 33.57 & 32.22 & 33.08 \\
 & SFT & \cellcolor{depthGain!23.43}48.12 & \cellcolor{depthGain!25.83}42.65 & \cellcolor{depthGain!29.93}41.44 & \cellcolor{depthGain!34.21}41.40 & \cellcolor{depthGain!31.78}41.56 \\
 & \method{} & \cellcolor{depthGain!31.92}52.27 & \cellcolor{depthGain!32.71}45.56 & \cellcolor{depthGain!39.72}45.31 & \cellcolor{depthGain!40.79}43.89 & \cellcolor{depthGain!37.34}43.72 \\
\bottomrule
\end{tabular}
\vspace{2mm}
\end{table}

\method{} exceeds Base at every tested depth in all four pooled paths.
Pooling the four paths yields gains of 9.61--11.74~pp over Base and
2.16--3.86~pp over SFT at $k=1$--$4$.

%% file: tables/supplementary_models.tex
\begin{table}[htbp!]
\caption{Full results for Qwen3-4B-Instruct-2507 and Qwen2.5-7B-Instruct under the same conditions as Table~\ref{tab:main_results}.}
\label{tab:supplementary_models}
\centering
\definecolor{resultLoss}{HTML}{e66565}
\definecolor{resultGain}{HTML}{90EE90}
\definecolor{resultTie}{HTML}{FFFFE0}
\setlength{\tabcolsep}{3.6pt}
\renewcommand{\arraystretch}{1.2}
\resizebox{0.95\linewidth}{!}{
\begin{tabular}{@{}llrrrrrrrr@{}}
\toprule
\multirow{2}{*}{\textbf{Model}} & \multirow{2}{*}{\textbf{Domain}}
& \multicolumn{4}{c}{\textbf{Single}}
& \multicolumn{4}{c}{\textbf{Multi}} \\
\cmidrule(lr){3-6}\cmidrule(lr){7-10}
& & Original & Direct Rev. & Revised & Retained
& Original & Neutral & Revised & Retained \\
\midrule
\multirow{4}{*}{\qwen~3-4B}
& \domainlabel{Math}    & 86.41 & \cellcolor{resultGain!10.95}87.38 & \cellcolor{resultLoss!15.73}80.58 & \cellcolor{resultLoss!19.55}76.70 & \cellcolor{resultLoss!41.52}54.37 & \cellcolor{resultLoss!42.47}53.40 & \cellcolor{resultLoss!45.33}50.49 & \cellcolor{resultLoss!68.26}27.18 \\
& \domainlabel{Code}    & 78.00 & \cellcolor{resultGain!11.09}79.00 & \cellcolor{resultLoss!49.23}42.00 & \cellcolor{resultLoss!11.09}77.00 & \cellcolor{resultLoss!64.49}28.00 & \cellcolor{resultLoss!62.31}30.00 & \cellcolor{resultLoss!64.49}28.00 & \cellcolor{resultLoss!68.85}24.00 \\
& \domainlabel{Database} & 90.65 & \cellcolor{resultLoss!10.87}89.72 & \cellcolor{resultLoss!24.89}74.77 & \cellcolor{resultLoss!30.15}69.16 & \cellcolor{resultLoss!59.95}37.38 & \cellcolor{resultLoss!68.71}28.04 & \cellcolor{resultLoss!59.95}37.38 & \cellcolor{resultLoss!73.97}22.43 \\
& \domainlabel{Actions} & 92.31 & \cellcolor{resultLoss!10.88}91.35 & \cellcolor{resultLoss!24.17}76.92 & \cellcolor{resultLoss!17.08}84.62 & \cellcolor{resultLoss!50.73}48.08 & \cellcolor{resultLoss!50.73}48.08 & \cellcolor{resultLoss!55.16}43.27 & \cellcolor{resultLoss!63.12}34.62 \\
\midrule
\multirow{4}{*}{\qwen~2.5-7B}
& \domainlabel{Math}    & 89.32 & \cellcolor{resultLoss!11.85}87.38 & \cellcolor{resultLoss!24.78}73.79 & \cellcolor{resultLoss!34.94}63.11 & \cellcolor{resultLoss!47.88}49.51 & \cellcolor{resultLoss!48.80}48.54 & \cellcolor{resultLoss!50.65}46.60 & \cellcolor{resultLoss!75.00}20.39 \\
& \domainlabel{Code}    & 68.00 & \cellcolor{resultGain!11.25}69.00 & \cellcolor{resultLoss!17.50}62.00 & \cellcolor{resultLoss!41.25}43.00 & \cellcolor{resultLoss!45.00}40.00 & \cellcolor{resultLoss!40.00}44.00 & \cellcolor{resultLoss!52.50}34.00 & \cellcolor{resultLoss!61.25}27.00 \\
& \domainlabel{Database} & 81.31 & \cellcolor{resultLoss!11.95}79.44 & \cellcolor{resultLoss!16.84}74.77 & \cellcolor{resultLoss!52.99}40.19 & \cellcolor{resultLoss!52.99}40.19 & \cellcolor{resultLoss!58.85}34.58 & \cellcolor{resultLoss!59.83}33.64 & \cellcolor{resultLoss!74.48}19.63 \\
& \domainlabel{Actions} & 97.12 & \cellcolor{resultLoss!12.53}94.23 & \cellcolor{resultLoss!31.88}72.12 & \cellcolor{resultLoss!29.36}75.00 & \cellcolor{resultLoss!53.76}47.12 & \cellcolor{resultLoss!51.24}50.00 & \cellcolor{resultLoss!52.08}49.04 & \cellcolor{resultLoss!61.34}38.46 \\
\bottomrule
\end{tabular}
}
\end{table}

%% file: appendix/f_discussion.tex
\section{Limitations and Discussion}
\label{app:limitations_discussion}

\textit{\textbf{Interaction coverage.}}
\benchmark{} uses controlled clarification, rejection, and revision events
to compare responses to the same source tasks under different interaction
conditions. This design makes the active requirements and reference
answers explicit and verifiable. The current benchmark focuses on
decisions about one selected requirement. Extending coverage to multiple
interacting requirements and more varied user behavior remains future work.

\textit{\textbf{Method development.}}
\method{} provides a simple training approach to help LLMs identify
and follow active intent in multi-turn dialogue. It improves overall
accuracy over Base and matched SFT in our experiments. However, mean \textit{Retained} accuracy is similar to SFT.
Handling rejected changes remains challenging and is an important
direction for further method development. 

\textit{\textbf{Supervision from natural conversations.}}
Our training uses complete active-task specifications supplied by the
benchmark construction. These specifications support the Teacher during
training, while the trained Student uses only the dialogue at inference.
Applying the approach to natural conversations requires reconstructing
active requirements from user messages and decisions. Future work could
build these specifications by extracting and updating needs
from dialogue, with consistency checks and human review for ambiguous cases.